\documentclass{bmvc2k}

\usepackage{comment}
\usepackage{array}
\newcommand{\PreserveBackslash}[1]{\let\temp=\\#1\let\\=\temp}
\newcolumntype{C}[1]{>{\PreserveBackslash\centering}p{#1}}
\usepackage{amssymb}
\usepackage{wrapfig}
\usepackage{algorithm}
\usepackage{algorithmicx}
\usepackage{algpseudocode}

\title{Beyond deterministic translation for unsupervised domain adaptation}

\addauthor{Eleni Chiou}{eleni.chiou.17@ucl.ac.uk}{1}
\addauthor{Eleftheria Panagiotaki}{e.panagiotaki@ucl.ac.uk}{1}
\addauthor{Iasonas Kokkinos}{i.kokkinos@cs.ucl.ac.uk}{1,2}

\addinstitution{
 Department of Computer Science \\
 University College London \\
 London, UK
}

\addinstitution{
 Snap Inc.
}

\runninghead{Chiou, Panagiotaki, Kokkinos}{Beyond deterministic translation for UDA}

\begin{document}

\maketitle

\begin{abstract}
In this work we challenge the common approach  of using a one-to-one mapping (`translation')  between the source and target domains in 
unsupervised domain adaptation (UDA). 
Instead, we rely on \textit{stochastic translation} to capture inherent translation ambiguities. This allows us to (i) train more accurate target networks by generating multiple outputs conditioned on the same source image,  leveraging both accurate translation and data augmentation for appearance variability, 
(ii) impute robust pseudo-labels for the target data by averaging the predictions of a source network on multiple translated versions of a single target image 
and (iii) train and ensemble diverse networks in the target domain by modulating the degree of stochasticity in the translations. 
We report improvements over strong recent baselines, leading to
state-of-the-art UDA results on two challenging semantic segmentation benchmarks. Our code is available at \scriptsize{\url{https://github.com/elchiou/Beyond-deterministic-translation-for-UDA}}.
\end{abstract}
\vspace{-.48cm}
\begin{figure}[!ht]
\centering
\includegraphics[width=0.8\textwidth]{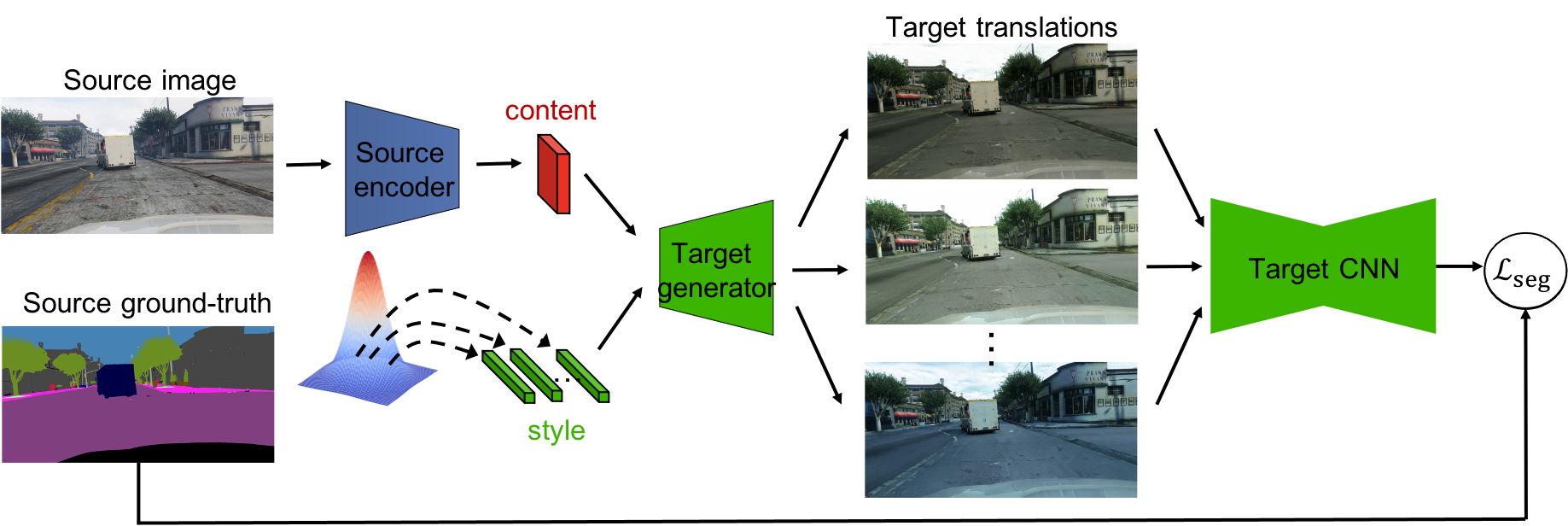}
\vspace{-.20cm}
    \caption{Unsupervised Domain Adaptation (UDA) with stochastic translation: we rely on a content-style separation network to associate a synthetic image from the GTA5 dataset (source) with a distribution of image translations to the target domain. These translations preserve the content signal and adopt the appearance properties of the Cityscapes dataset (target). We use the resulting images to train a target-domain network tasked with predicting the labels of the respective source-domain image, irrespective of the style variation.
}
\label{fig:stoch_transl}
\end{figure}

\vspace{-0.80cm}

\section{Introduction}
Unsupervised Domain Adaptation (UDA) aims at accommodating the differing statistics between a `source' and a `target' domain, where the source domain comes with input-label pairs for a task, while the target domain only contains input samples. Successfully solving this problem can allow us for instance  to exploit  synthetically generated datasets that come with rich ground-truth to train models that can perform well in real images with different appearance properties. Translation-based approaches ~\cite{Hoffman_18_cycada,Li_19_BDL,yang_20_PCEDA,Cheng_21_dual_path,yang_2020_label_driv_rec} 
rely on establishing a transformation between the two domains (often referred to as `pixel space alignment') that bridges the difference in their statistics while preserving the semantics of the translated samples. This translation can then be used as a mechanism for generating supervision in the `target' domain based on ground-truth originally available in a `source' domain.

In this work we address a major shortcoming of this approach - namely the assumption that this translation is a deterministic function, mapping a single source  to a single target image. Recent works on the closely related problem of unsupervised image translation~\cite{Huang_18_munit,Almahairi_18a_augm_cycleGAN,Lee_18_DRIT,Zhu_17_BicycleGAN} have highlighted that this is a strong assumption and is frequently violated in practice.  For instance a nighttime scene can have multiple daytime counterparts where originally invisible structures are revealed by the sun and also illuminated from different directions during the day. To mitigate this problem these techniques introduce methods for multimodal, or stochastic translation, that allows an image from one domain to be associated with a whole distribution of images in another. An earlier work~\cite{Chiou_20_harn_unc} has shown the potential of generating multiple translations in the narrow setting of supervised domain adaptation across different medical imaging modalities. In this work we exploit stochasticity in the problem of UDA in three complementary ways and show that stochastic translation improves upon the current state-of-the-art in UDA on challenging semantic segmentation benchmarks.

Firstly, we use stochastic  translation across the source and target domains by relying on the multimodal (or stochastic) translation method of~\cite{Huang_18_munit}.
We show that allowing for stochastic translations yields clear improvements over the deterministic CycleGAN-based baseline, as well as all published pixel space alignment-based techniques. We attribute this to the ability of the multimodal translation to generate more diverse and sharper samples, that provide better training signals to the target-domain network. 

Secondly, we exploit the ability to sample multiple translations for a given image in order to obtain better pseudo-labels for the unlabelled target images: we generate multiple translations of every target image into the source domain, label each according to a source-domain CNN, and average the resulting predictions to form a reliable estimate of the class probability. This is used as supervision for target-domain networks, and is shown to be increasingly useful as the number of averaged samples per image grows. 

Thirdly, we modify the variance of the latent style code in order to train and ensemble complementary target-domain networks, each of which is adapted to handle a different degree of appearance variability. The results of ensembling these networks on the target data are then used to train a single target-domain network that outperforms all methods that also rely on ensembling-based supervision in the target domain.


We show that each of our proposed contributions yields additional improvements over strong recent baselines, leading to state-of-the-art UDA results on two challenging semantic segmentation benchmarks. 


\vspace{-.30cm}

\section{Related Work}
UDA approaches aim at learning domain invariant representations by aligning the distributions of the two domains at feature/output level ~\cite{tsai_2018_AdaptSegNet,wang_2020_classes_matter,zhang_2019_category_anchor,ma_2021_coarse_to_fine,mei_2020_instance,chang_19_all_about_structure,pan_2020_intra_domain} or at image level~\cite{Hoffman_18_cycada,Li_19_BDL,Cheng_21_dual_path}. Based on the observation that the source and the target domain share a similar semantic layout,~\cite{tsai_2018_AdaptSegNet,vu_2019_advent} rely on adversarial training to align the raw output and entropy distributions respectively. However, such a global alignment  does not guarantee that individual target samples are correctly classified. Category-based feature alignment methods ~\cite{saito_2018_maximum_classifier,tsai_2019_discriminative_patch,luo_2019_taking_a_closer,Wang_2020_diff_treat,wang_2020_classes_matter,zhang_2019_category_anchor} attempt to address this problem by mapping target-domain features closer to the corresponding source-domain features. 

Image-level UDA methods aim at aligning the two domain at the raw pixel space.~\cite{Hoffman_18_cycada,Li_19_BDL,yang_20_PCEDA,Cheng_21_dual_path} rely on  CycleGAN~\cite{Zhu_2017_CycleGAN} to translate source domain images to the style of the target domain. Two recent works~\cite{yang_2020_FDA,ma_2021_coarse_to_fine} bypass the need for training an image translation network by relying on simple Fourier transform and global photometric alignment respectively.

Complementary to the idea of translation is the use of self-training ~\cite{zou_2018_class_balanced_self_training,zou_2019_confidence_regularized,two_phase_ps,zhang_2021_proda}  which has been originally used in semi-supervised learning. Self-training iteratively generates pseudo-labels for the target domain based on confident predictions and uses those to supervise the model, implicitly encouraging category-based feature alignment between the source and the target domain. Another direction pursued in~\cite{choi_2019_self_ensembiling,melas_2021_pixmatch} is to leverage the unlabeled target data by using consistency regularization to make the model predictions invariant to perturbations imposed in the input images. 


Two recent works~\cite{Li_19_BDL,Cheng_21_dual_path} that rely on both image-level alignment and self-training are more closely related to our work.~\cite{Li_19_BDL} relies on CycleGan to translate source images to the style of the target domain. They train the image translation network and the segmentation network alternatively and introduce a perceptual supervision based on the segmentation network to enforce semantic consistency during translation. They also  generate pseudo-labels for the target data based on high confident predictions of the target network and use those to supervise the target network. \cite{Cheng_21_dual_path} improves upon~\cite{Li_19_BDL} by replacing the single-domain perceptual supervision with a cross-domain perceptual supervision using two segmentation networks operating in the source and the target domain respectively. In addition, they rely on both the source and the target networks to generate pseudo labels for the target data. Similar to these works we rely on image-to-image translation to translate source images to the style of the target domain, but we go beyond their one-to-one mapping approach which allows us to leverage both accurate translation and data augmentation for appearance variability. In addition, as in \cite{Cheng_21_dual_path} we  use  source and target networks to generate pseudo-labels, but we exploit stochasticity in the translation to generate more robust pseudo-labels that allow us to train accurate target-domain networks. 

\newcommand{\ba}{\begin{eqnarray*}}
\newcommand{\ea}{\end{eqnarray*}}
\newcommand{\beq}{\begin{equation}}
\newcommand{\eeq}{\end{equation}}
\newcommand{\loss}{\mathcal{L}}
\newcommand{\refeq}[1]{Eq.~\ref{#1}}
\newcommand{\transsym}{\mathbf{T}}
\newcommand{\trans}[1]{\mathbf{T}[#1]}
\newcommand{\transd}[2]{\mathbf{T}[#1,#2]}
\newcommand{\noise}{\mathbf{v}}
\newcommand{\normal}{\mathcal{N}(\mathbf{0},\mathbf{I})}
\newcommand{\reffig}[1]{Fig.~\ref{#1}}
\newcommand{\refsec}[1]{Sec.~\ref{#1}}
\newcommand{\bal}{\begin{aligned}}
\newcommand{\eal}{\end{aligned}}

\vspace{-.30cm}

\section{Methods}
We start in \refsec{sec:intro} by introducing the background of
using translation in UDA, and
then introduce our technical contributions from \refsec{sec:stochastic}
onwards. 
\vspace{-.30cm}
\subsection{Domain Translation and UDA}
\label{sec:intro}
In UDA we consider a source dataset with paired image-label data:
$\mathcal{S} = \{(x^i_s,y^i_s)\}, i \in [1,S]$ and a target dataset comprising only image data $\mathcal{T} = \{x^i_t\}, i \in [1,T]$. Our task is to learn a segmentation system that provides accurate predictions in the target domain; we assume a substantial domain gap, precluding the naive approach of training a network on $\mathcal{S}$ and then deploying it in the target domain.

Output-space alignment UDA approaches~\cite{tsai_2018_AdaptSegNet} train a single segmentation network, $F$ on both the source and the target images, using a cross-entropy loss in the source domain and an  adversarial loss in the target domain to  statistically align the predictions on target images to the distribution of source predictions. This  results in a training objective of the following form:
\begin{equation}
\loss(F) = \sum_{(x,y) \in \mathcal{S}} \loss_{ce}(F(x),y) + \sum_{x \in \mathcal{T}} \loss_{adv}(F(x)), \label{eq:adapt_seg}
\end{equation} 
where $F(x)$ the softmax output.

In \cite{vu_2019_advent}  entropy-based adversarial training is used to align the target entropy distribution to the source entropy distribution instead of aligning the raw predictions, resulting in the following objective:
\begin{equation}
\loss(F) = \sum_{(x,y) \in \mathcal{S}} \loss_{ce}(F(x),y) + \sum_{x \in \mathcal{T}} \loss_{adv}(E(F(x))), \label{eq:advent}
\end{equation} 
where $E(F(x)) = - F(x) \log(F(x))$ is the weighed self-information. 

Given that the network provides low-entropy predictions on source images, adversarial entropy minimization  promotes low-entropy predictions in the target domain.
Still, having a single network $F$ that successfully operates in both domains can be challenging due to the broader intra-class variability caused by the domain gap.

Pixel-space alignment approaches try to mitigate this problem by establishing a relation between the distributions of the source and target domain images and using that to supervise a network that only operates with target-domain images.
In its simplest form, adopted also in \cite{Bousmalis_17_pixel_adapt,Hoffman_18_cycada,wu2018dcan,Li_19_BDL,yang_20_PCEDA} this relation is a deterministic translation function $\transsym$ that maps source images to the target domain, resulting in the following objective:
\begin{equation}
\!\!\loss(F_t)  \!=\! \sum_{(x,y) \in \mathcal{S}} \loss_{ce}(F_t(\trans{x}),y)  +  \sum_{x \in \mathcal{T}} \loss_{adv}(E(F_t(x))), \label{eq:transladvent}
\end{equation} 
where the difference with respect to \refeq{eq:advent} is that the translated version of $x$, $\trans{x}$ is passed to the target-domain segmentation network, $F_t$.
A straightforward way of obtaining such a translation function is through unsupervised translation between the two domains \cite{Zhu_2017_CycleGAN}. 

This approach  creates a target-adapted variant of the source-domain dataset, allowing us to train a single network that is tuned exclusively to the statistics of the target domain. This reduces the intra-class variance and puts less strain on the segmentation network, but relies on the strong assumption that such a deterministic translation function exists. In this work we relax this assumption and work with a {\emph{ distribution on translated images}}. 
This better reflects most UDA scenarios and provides us with novel and simple tools to improve UDA performance, as described below. 

\vspace{-.30cm}
\subsection{Stochastic Translation and UDA}
\label{sec:stochastic}

We propose to replace the deterministic translation function $\trans{x}$, with a distribution over images given by  $\transd{x}{\noise}, \noise\sim\normal$, where $\noise$ is a random vector sampled from a  normal distribution with zero mean and unit covariance~\cite{Huang_18_munit}. For instance when translating a nighttime scene into its  daytime scene, the random argument can reflect the position of the sun, clouds or obscured objects. For the synthetic-to-real case that we handle in our experiments we can see from \reffig{fig:stoch_transl} that the translation network can indeed generate scenes illuminated differently as well as different cloud patterns, allowing us to capture more faithfully the range of scenes encountered in the target domain. 
We note that $\transsym$ remains deterministic and can be expressed by a neural network, but has a random argument which results in a distribution on translated images.

This change is reflected in the UDA training objective by replacing the loss of the translated image with the {\emph{ expected loss}} of the translated image:
\begin{equation}
\loss(F_t)  = \!\!\sum_{(x,y) \in \mathcal{S}} \mathbf{E}_{\noise} \left[\loss_{ce}(F_t(\transd{x}{\noise}),y)\right] + \sum_{x \in \mathcal{T}} \loss_{adv}(E(F_t(x))), \label{eq:stochtransladvent}
\end{equation}
 where  the expectation is taken with respect to the random vector $\noise \sim \normal$, driving the stochastic translation. We note that during training we create minibatches by first sampling images from $\mathcal{S}$ and then sampling $\noise$ once per image, effectively replacing the integration in the expectation with a Monte Carlo approximation.

\newcommand{\content}{C}
\newcommand{\style}{S}
\newcommand{\stylesample}{\mathbf{s}}

Our stochastic translation network relies on  MUNIT \cite{Huang_18_munit}: 
we start from reconstructing images in each domain through content and style encodings, where content is fed to the first layer of a generator whose subsequent layers are modulated by style-driven Adaptive Instance Normalization~\cite{huang2017arbitrary} - this amounts to minimizing the following domain-specific autoencoding objectives:
\ba
L_{s} = \sum_{x\in \mathcal{S}} \| x - G_{s}(\content_s(x),\style_s(x))\|, \nonumber &
L_{t} = \sum_{x\in \mathcal{T}} \|x - G_{t}(\content_t(x),\style_t(x))\|,
\ea
where $\content_s,\style_s,G_s$ are the content-encoder, style-encoder and generator networks for the source domain $s$ respectively, while $\content_t,\style_t, G_t$ are those of the target domain $t$.

The basic assumption is that the commonalities between two domains are captured by the shared content space, allowing us to pass content from the source image to its target counterparts, as also shown in \reffig{fig:stoch_transl}. The uncertainty in the translation is captured by a domain-specific style encoding that is inherently uncertain given the source image.

This results in the following stochastic translation  function from source to target:
\ba
\transd{x}{\noise} \doteq G_{t}(\content_s(x),\noise), ~ \noise \sim \normal,~x \in \mathcal{S},
\ea
where we encode the content of the source image through  $\content_s(x)$ and then pass it to the target-domain generator $G_{t}$ that is driven by a random style code $\noise$. A similar translation is established between the target and source domains, and adversarial losses on both domains ensure that the resulting translations appear as realistic samples of the respective domains.

The alignment of the shared latent space for content is enforced by a cycle translation objective:
\ba
L_{cycle}^c = \|\content_t(G_{t}(\content_s(x),\noise)) - \content_s(x)\|_2,~x \in \mathcal{S}, ~ \noise \sim \normal,
\ea
ensuring that regardless of the random style code, we can recover the original content $\content_s(x)$ by encoding the translated image through the respective content encoder. A similar loss is used for the style code:
\ba
L_{cycle}^s = \|\style_t(G_{t}(\content_s(x),\noise)) - \noise \|_2,~x \in \mathcal{S}, ~ \noise \sim \normal.
\ea

We preserve  semantic information  during translation by imposing a semantic consistency constraint to our stochastic translation  network using a fixed segmentation network $F$ pretrained on source and target data using~\refeq{eq:advent}. Given an image $x$ we obtain the predicted labels before translation as $p = \textrm{argmax}(F(x))$ and enforce semantic consistency during translation using an objective of the following form:
\begin{equation}
L_{sem} = \loss_{ce}(F(\transd{x}{\noise}), p).
\label{eq:semloss}
\end{equation}
The losses are applied to translations to both domains since unlike UDA, there is no special `source' and `target' domain. 

\begin{wrapfigure}{r}{0.45\textwidth}
\includegraphics[width=0.45\textwidth]{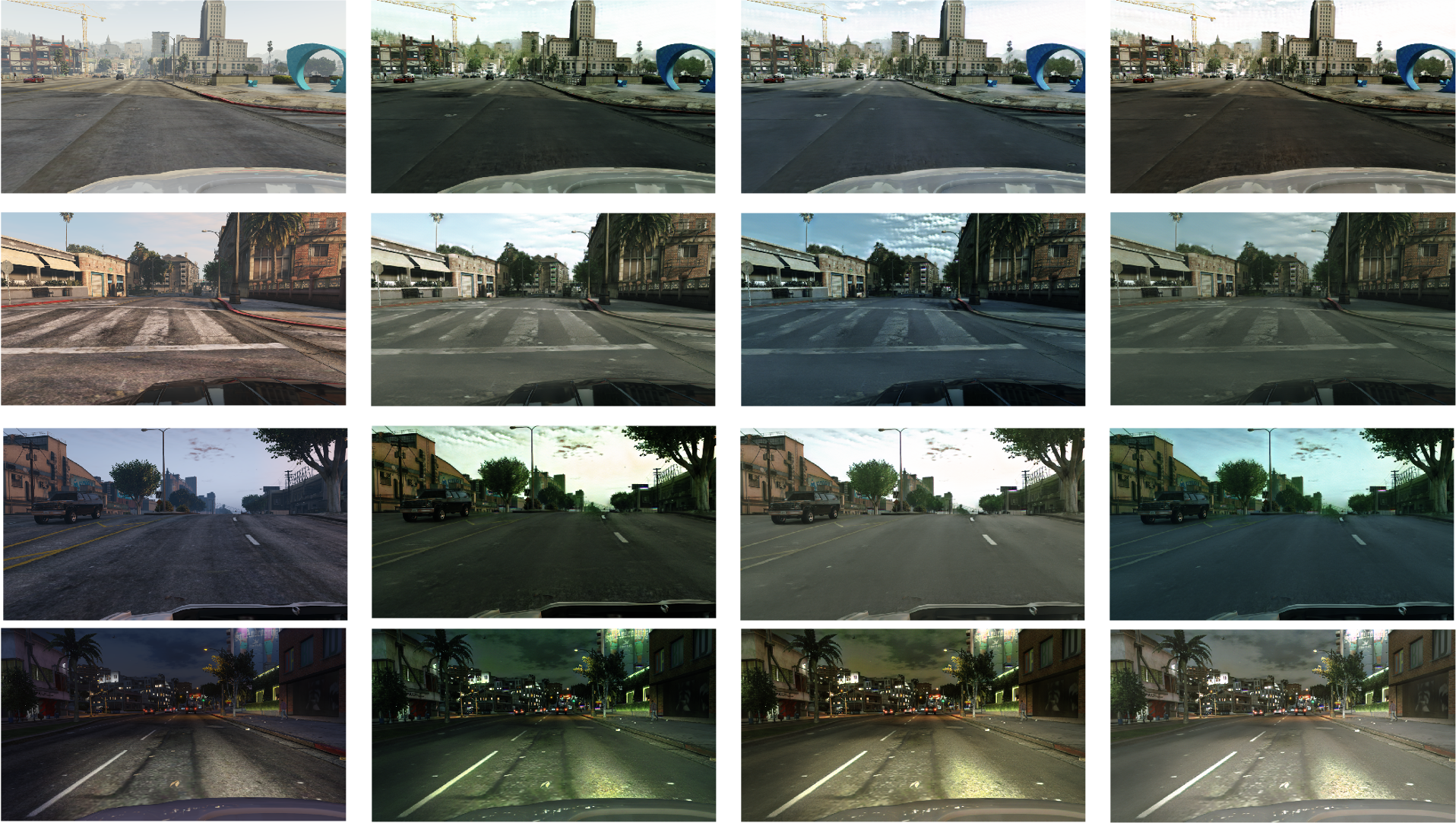}\\
\hspace{-.10\linewidth}Source \hspace{.18\linewidth} Target translations \\
\vspace{-.6cm}
\caption{Diverse translations of  images from the 
GTA 
source dataset to the Cityscapes target dataset. We generate diverse variants of the same scene, capturing more faithfully the data distribution in the target domain. 
}
\label{fig:diff_var}
\vspace{-.8cm}
\end{wrapfigure}
We argue that stochastic translation provides us with a natural mechanism to handle UDA problems with large domain gaps where things may unavoidably get `lost in translation'; the content cycle constraint can help preserve semantics during translation, while the random style allows the translated image appearance to vary freely, avoiding a deterministic and blunt translation. This is demonstrated in \reffig{fig:diff_var}, where we show some of the samples obtained by our method. 

\newcommand{\transidet}[1]{\mathbf{I}[#1]}
\newcommand{\transid}[2]{\mathbf{I}[#1,#2]}
\newcommand{\transi}{\mathbf{I}}

\vspace{-.30cm}
\subsection{Stochastic translation and pseudo-labelling}
\label{sec:pseudosource}
Having shown how stochastic translation from the source to the target domain can be integrated in the basic formulation of UDA, we now turn to exploiting stochastic translation from the target to the source domain, which is freely provided by the cycle-consistent formulation of \cite{Huang_18_munit}. 

In particular we consider a complementary segmentation network, $F_s$, that operates in the source domain and can be directly supervised from the labeled source dataset based on a cross-entropy objective: 
\begin{equation}
\loss(F_s) = \sum_{(x,y) \in \mathcal{S}} \loss_{ce}(F_s(x),y).  \label{eq:source_seg}
\end{equation}
This network can provide labels for the target-domain images, once these are translated from the target to the source domain; these pseudo-labels of the target data can in turn be used to supervise the target-domain network through a cross-entropy loss. 
\begin{wrapfigure}{r}{0.5\textwidth}
\vspace*{-.4cm}
\includegraphics[width=0.5\textwidth]{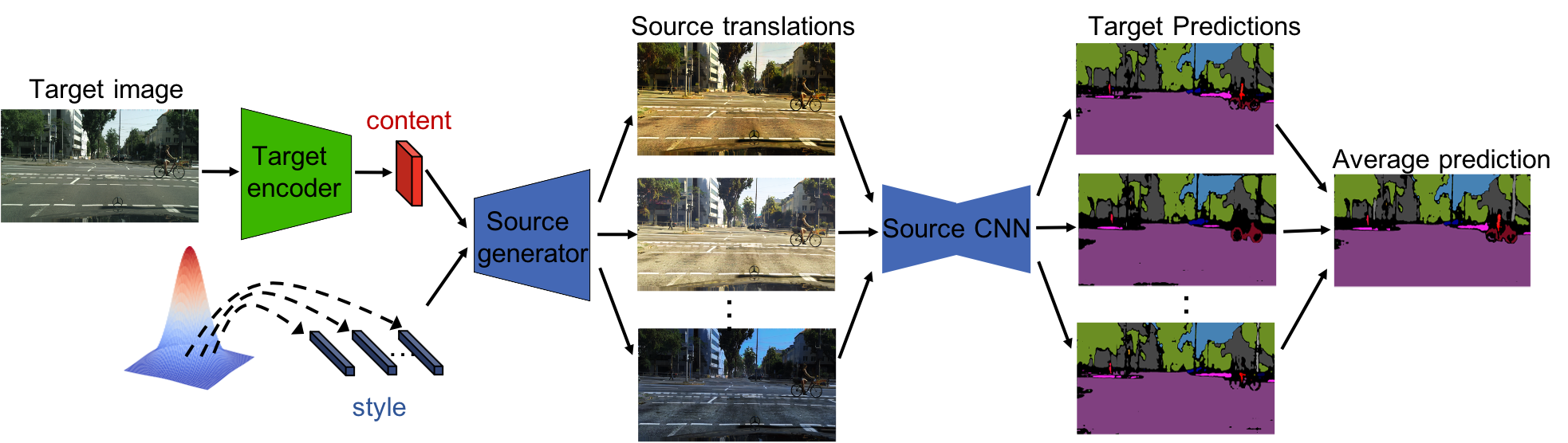}
\vspace*{-.6cm}
\caption{Stochastic translation for pseudo-labeling: the target image (left) results in multiple source-domain translations which are processed by the source-domain network, $F_s$ and averaged to produce pseudo-labels. 
}
\label{fig:pseudo_lab}
\vspace*{-.2cm}
\end{wrapfigure}
In the case of deterministic translation pseudo-labels would be obtained by the following expression:
\begin{equation}
\hat{y}(x) = F_s(\transidet{x}), \quad x \in \mathcal{T}, \label{eq:pseudo}
\end{equation}
where $\transi$ is the inverse transform from the target to the source domain, and $\hat{y}$ indicates the pixel-level posterior distribution on labels. 

In our case however we have a whole distribution on translations for every image in $\mathcal{T}$. We realise that we can exploit multiple samples to obtain a better estimate of the pseudo-labels. In particular we form the following Monte Carlo estimate of pseudo-labels: 
\ba
\hat{y}(x) &=& E_{\noise} \left[F_s(\transid{x}{\noise})\right], \quad x \in \mathcal{T}, \noise \sim \normal  \\
&\simeq& \frac{1}{K}\sum_{k=1}^{K}F_s(\transid{x}{\noise_k}), 
\label{eq:pseudoMC}
\ea
where $\noise_k$ are  independently sampled from the normal distribution. As shown in  \reffig{fig:pseudo_lab} the label maps obtained through this process tend to have fewer errors and be more confident, since averaging the results obtained by different translations can be expected to cancel out the fluctuation of the predictions around their ground-truth value. 

Our experimental results indicate that using $K=10$ yields substantially better results than using a single sample. We also note that pseudo-label generation is a one-off process done prior to training the target-domain network, and consequently the number of samples, $K$, does not affect training time. 

\vspace{-.30cm}
\subsection{Stochasticity-driven training of diverse network ensembles}
\label{sec:ensembling}

An experimental approach that has been recently adopted by several recent works \cite{yang_2020_FDA,Cheng_21_dual_path} consists in ensembling different networks trained for UDA, and using their predictions as an enhanced pseudo-labeling mechanism. 

Based on the understanding that the stochasticity driving our translation mechanism can be seen as implementing appearance-level dataset augmentation in the target domain, we 
introduce a simple twist to the translation mechanism that allows us to train networks that operate in different regimes.
For this we scale by a constant the variance of the normal distribution used to sample the random style code - this amounts to generating more diverse translations than those suggested by 
the image statistics of the target domain. On one hand this trains a target network that can handle a broader range of inputs, but on the other hand it may  waste capacity to handle
unrepresentative samples. 

\begin{wrapfigure}{r}{0.5\textwidth}
\vspace{-.5cm}
\centering
\includegraphics[width=0.5\textwidth]{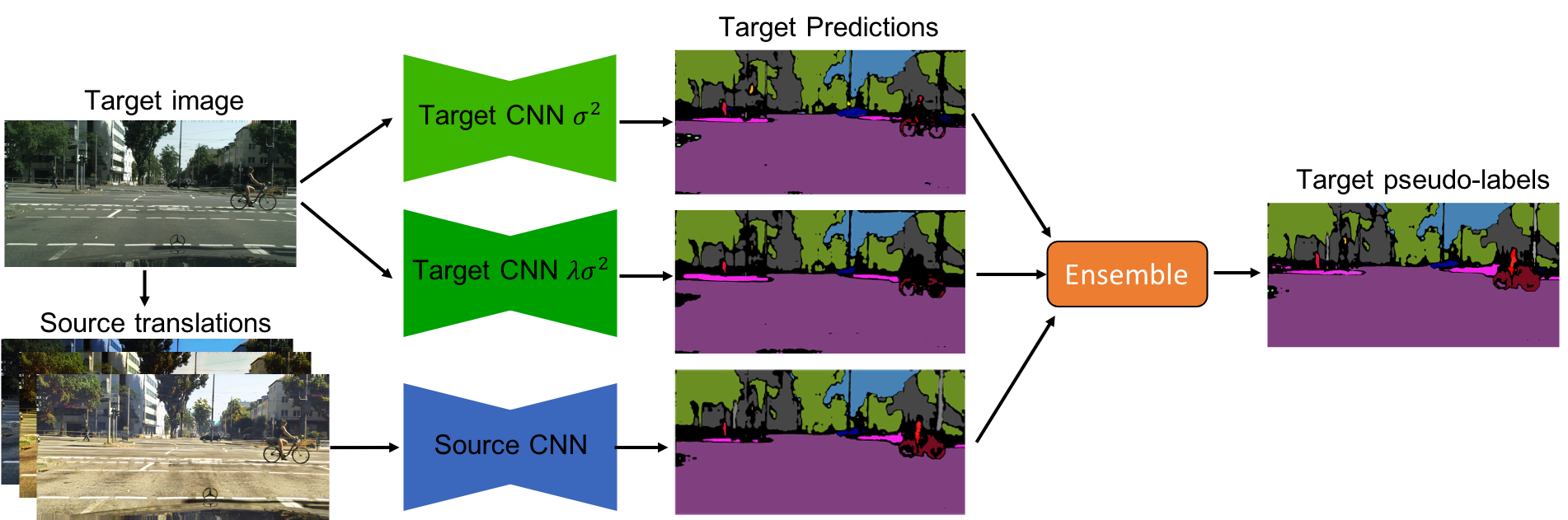}
\vspace{-.4cm}
\caption{Ensembling of a triplet of networks  ---  two target networks trained with different degrees of stochasticity in the translation ($\sigma^2$) and a source network --- for robust pseudo-labeling. 
}
\vspace{-.4cm}
\label{fig:ensemble}
\end{wrapfigure}

We train two such networks, one with the variance left intact and the other with the variance scaled by 10, and average their predictions with those of the source-domain network described in the previous subsection as shown in \reffig{fig:ensemble}.
Our results show that this triplet of networks yields a clear boost over the baseline operating with a single network. 

Further following common practice in UDA we use the resulting ensembling results as pseudo-labels in the next round of training - this yields further improvements, as documented in detail in the experimental results section. 

\vspace{-.30cm}
\subsection{Training objectives}
\label{sec:learning}
Firstly, we train our stochastic translation network using the  process of \cite{Huang_18_munit} and introduce a semantic consistency loss as in \cite{Hoffman_18_cycada} to preserve semantics during translation.

For the target-domain network the basic objective has already been provided in \refeq{eq:stochtransladvent},
where $\loss_{ce}$ is the standard cross-entropy loss and $\loss_{adv}$ is the adversarial entropy minimization objective \cite{vu_2019_advent}. A more sophisticated objective can train this network  with pseudo-labels, obtained either from a source-domain network as described in \refsec{sec:pseudosource} or from the ensembling  of multiple  networks, as described in  \refsec{sec:ensembling}. In that case the objective becomes:
\begin{equation}
\begin{gathered}
\loss(F_t) = \sum_{(x,y) \in \mathcal{S}} \mathbf{E}_{\noise} \left[\loss_{ce}(F_t(\transd{x}{\noise}),y)\right] + \\
\sum_{x \in \mathcal{T}} \loss_{adv}(E(F_t(x))) + \sum_{x \in \mathcal{T}} \loss_{ce}^{\theta}(F_t(x),\textrm{argmax}(\hat{y}) ), \label{eq:distillation}
\end{gathered}
\end{equation}
where the cross entropy loss $\loss_{ce}^{\theta}(F_t(x))$ is only applied on pseudo-labels where the dominant class has a score above a certain threshold $\theta$.
Similar to~\cite{zou_2018_class_balanced_self_training} we use class-wise confidence thresholds to address the inter-class imbalance and avoid ignoring hard classes. 
We provide  more details in the supplementary material. 

Finally, for the source-domain network, we observed experimentally that we obtain better results by adding an entropy-based regularization to the output of $F_s$ when it is driven by translated target images
The objective function for the source network becomes: 
\begin{equation}
\loss(F_s) = \sum_{(x,y) \in \mathcal{S}} \loss_{ce}(F_s(x),y) 
 + \sum_{x \in \mathcal{T}} \mathbf{E}_{\noise} \left[\loss_{adv}(F_s(\transid{x}{\noise}))\right],
\label{eq:stochtransladventsource}
\end{equation} 
forming the source-domain counterpart to the objective encountered in \refeq{eq:stochtransladvent}. When pseudo-labels are available for the target domain, we train the source network using 
the source-domain counterpart of the objective in \refeq{eq:distillation}.

\section{Experiments}
We evaluate the proposed approach on two common UDA benchmarks for semantic segmentation. In particular we use the synthetic dataset GTA5~\cite{Richter_2016_GTA5} or SYNTHIA~\cite{Ros_2016_SYNTHIA} with ground-truth annotations as the source domain and the Cityscapes~\cite{Cordts_2016_CitySc} dataset as the target domain with no available annotations during training. We provide details about the datasets in the supplementary material. We evaluate the performance using the mean intersection over union score (mIoU) across semantic classes on the Cityscapes validation set. We rely on MUNIT~\cite{Huang_18_munit} to establish a stochastic translation across the source and target domain. We train two different architectures, i.e., DeepLabV2 \cite{chen_2017_deeplab} with  ResNet101~\cite{he_2016_resnet} backbone, and FCN-8s~\cite{long_2015_fully}  with VGG-16~\cite{simonyan_2014_VGG} backbone. We provide implementation details in the supplementary material.

\textbf{Stochastic translation:} We start by examining in how stochastic translation improves performance compared to deterministic translation. In all cases the segmentation model is DeepLabV2~\cite{chen_2017_deeplab} and the source and target datasets are GTA5~\cite{Richter_2016_GTA5} and Cityscapes~\cite{Cordts_2016_CitySc} respectively. In Table~\ref{tab:stoch_gta2city} we start with an apples-to-apples comparison that builds on directly on the ADVENT baseline \cite{vu_2019_advent}; the first two rows compare the originally published  and our reproduced numbers respectively. The third row shows the substantial improvement attained by training the system of ADVENT using translated images - which amount to training with \refeq{eq:transladvent}. The forth row reports our stochastic translation-based result, amounting to training with \refeq{eq:stochtransladvent}. We observe a substantial improvement, that can be attributed solely to the stochasticity of the translation. The last row shows that imposing a semantic consistency constraint as described in~\refeq{eq:semloss} further improves the performance.

\begin{table}
\parbox{.5\linewidth}{
\centering
\resizebox{\linewidth}{!}{
\begin{tabular}{cccc} \\ \hline
{Method}                                       & {Output space}   & {Pixel space}              & {mIoU} \\  \hline
ADVENT \cite{vu_2019_advent}                   & \checkmark       &                            &  43.8   \\ \hline
ADVENT $^{\ast}$                               &  \checkmark      &                            &  42.9   \\ \hline
ADVENT $^{\ast}+$ \\ CycleGAN$^{\ast}$           &  \checkmark      & \checkmark                 &  45.1 \\ \hline
Ours                                                              &  \checkmark                & \checkmark &  46.2  \\ \hline
Ours w/ $L_{sem}$                                                 &  \checkmark  & \checkmark &  \textbf{46.6} \\ \hline
\end{tabular}}
\vspace{0.001cm}
\caption{GTA to Cityscapes UDA using stochastic translation: We train ADVENT using synthetic images obtained from deterministic translation (CycleGAN) and stochastic translation (Ours). We observe a clear improvement
thanks to pixel-space alignment based on stochastic translation. $^{\ast}$ denotes our retrained models}
\label{tab:stoch_gta2city}}
\hspace{2mm}
\parbox{.48\linewidth}{
\centering
\resizebox{\linewidth}{!}{
\begin{tabular}{|C{1.6cm}|C{1.6cm}|C{1.6cm}|C{1.6cm}|C{1.8cm}|C{1.6cm}|}  \hline
{$F_s$, K=1} & {$F_s$, K=5}                & {$F_s$, K=10}            & {$F_t$, $\sigma^2=1$}  & {$F_t$, $\sigma^2=10$} & {mIoU}         \\ \hline
\checkmark       &                         &                          &                        &                         &  43.3          \\ \hline
                 & \checkmark              &                          &                        &                         &  44.0          \\ \hline
                 &                         & \checkmark               &                        &                         &  44.4          \\ \hline
                 &                         &                          & \checkmark             &                         &  46.6          \\ \hline
                 &                         &                          &                        & \checkmark              &  46.1          \\ \hline
                 &                         & \checkmark               & \checkmark             &                         &  47.7          \\ \hline
                 &                         & \checkmark               &                        & \checkmark              &  47.6          \\ \hline
                 &                         &                          & \checkmark             & \checkmark              &  47.7          \\ \hline
                 &                         & \checkmark               & \checkmark             & \checkmark              &  \textbf{48.2} \\ \hline
\end{tabular}}
\vspace{0.001cm}
\caption{Performance of different models and their combinations. The first 3 rows show the performance of the source network $F_s$ when averaging the predictions of multiple translations $K$, of a  target image while rows 4, 5 show the performance of the target networks $F_t$, trained with different degrees of stochasticity ($\sigma^2$) in the translation. Averaging the predictions of multiple translations and combining the three models allows us to obtain better pseudo-labels for the target domain.}
\label{tab:ens_gta_cityscapes}}
\vspace{-0.7cm}
\end{table}

\noindent\textbf{Pseudo labeling}:
As  discussed in Section \ref{sec:pseudosource} we translate from the target to the source domain  and generate pseudo labels for the target data.  The first three rows in Table~\ref{tab:ens_gta_cityscapes} show the impact of the number of samples $K$, on performance. Averaging the predictions of multiple translations for a given target image improves the performance and allows to obtain better pseudo labels for the target domain. Our results show that using 10 samples yields better performance. In rows 4, 5 of the same table we report the performance obtained from the two target networks trained with different degrees of stochasticity in the translation as described in \refsec{sec:ensembling}. Averaging the prediction of the three networks gives the best results, indicating the complementary of the model predictions. We also provide qualitative results in the supplementary material. 

\noindent\textbf{Network ensembling}: 
Table~\ref{tab:ens_gta_cityscapes_2_iters} shows the results obtained in three rounds of pseudo-labeling and training, following the approach of \cite{yang_2020_FDA,Li_19_BDL,Cheng_21_dual_path}. In the first round ($R=0$) we train the target and source networks with \refeq{eq:stochtransladvent} and \refeq{eq:stochtransladventsource} respectively using the synthetic and real data and  average the predictions of the three models to generate pseudo-labels for the target data.
\begin{wraptable}{r}{0.3\textwidth}
\vspace{-1.1cm}
\begin{center}
\resizebox{1\linewidth}{!}{
            \begin{tabular}{l c} \\ \hline
                 {Model} & {mIoU} \\  \hline

                $F_s$ (R=0) 	            & 44.4 \\

                $F_t$, $\sigma^2=1$ (R=0)  & 46.6  \\

                $F_t$, $\sigma^2=10$ (R=0) & 46.1 \\

                Ens (R=0)    & 48.2  \\ \hline

                $F_s$ (R=1) & 49.1  \\

                $F_t$, $\sigma^2=1$ (R=1) & 50.1 \\

                $F_t$, $\sigma^2=10$ (R=1) & 50.9 \\
                
                Ens (R=1) & 52.0   \\ \hline
                 
                $F_s$ (R=2)  &  51.3  \\

                $F_t$, $\sigma^2=1$ (R=2)  &  53.0 \\

                $F_t$, $\sigma^2=10$ (R=2) &  52.9 \\
                
                Ens (R=2) &    54.3   \\  \hline
            \end{tabular}}
        \end{center}
        \vspace{-0.2cm}
     \caption{Ablation study on GTA to Cityscapes. Averaging the predictions (Ens) of a source network $F_s$, and two target networks $F_t$ trained with different degrees of stochasticity ($\sigma^2$) in the translation allows to obtain robust pseudo-labels, while using multiple rounds $R$ of pseudo-labeling and training improves the overall performance.} \label{tab:ens_gta_cityscapes_2_iters}
    \vspace{-0.8cm}
\end{wraptable}
In the second round (R=1) we use the generated pseudo-labels as ground-truth labels to train the target and source networks with \refeq{eq:distillation} and its source-domain counterpart respectively. We observe that the pseudo-labels obtained by ensembling improve the performance of each individual network, as well as the ensemble obtained in the last round (R=2).

\noindent \textbf{Benchmark results}: We use DeepLabV2~\cite{chen_2017_deeplab}  with  ResNet101~\cite{he_2016_resnet} backbone, and FCN-8s~\cite{long_2015_fully}  with VGG-16~\cite{simonyan_2014_VGG} for the segmentation and compare with~\cite{vu_2019_advent,Li_19_BDL,yang_2020_FDA,yang_20_PCEDA,melas_2021_pixmatch,Cheng_21_dual_path} which use exactly the same experimental settings. We report both the results obtained using a single target network and the results obtained by ensembling. We provide qualitative results in the supplementary material.
The results (per-class IoU and mIoU over 19 classes) for the \textbf{GTA-to-Cityscapes} benchmark and ResNet101 backbone  are summarized in Table~\ref{tab:benchmark_gta2city}. In the supplementary material we provide results obtained using FCN-8s with VGG-16 and comparison with additional state-of-the-art methods. Our results show that our method achieves state-of-the-art performance and outperforms previous methods. When compared with other approaches relying on both deterministic translation and multiple rounds of pseudo-labeling and training~\cite{Li_19_BDL,Cheng_21_dual_path,yang_2020_FDA}, our approach performs better while at the same time is simpler. 
The results for the \textbf{SYNTHIA-to-Cityscapes} benchmark and  ResNet101 backbone are reported in Table~\ref{tab:benchmark_syn2city}. In the supplementary material we provide results obtained using FCN-8s with VGG-16 and comparison with additional state-of-the-art methods. Following the evaluation protocol of previous studies~\cite{Li_19_BDL,yang_20_PCEDA,yang_2020_FDA,vu_2019_advent,Cheng_21_dual_path} we report the mIoU of our method on $13$ and $16$ classes. We observe that our method outperforms previous state-of-the art methods by a large margin (+3.7 compared to DPL\cite{Cheng_21_dual_path}). We note here that the domain gap between SYNTHIA and Cityscapes is much larger compared to the domain gap between GTA and Cityscapes. We attribute the substantial improvements obtained by our method to the stochasticity in the translation which allows us to better capture  the range of scenes encountered in the two domains and to generate sharp samples even in cases where there is a large domain gap between the two domains. 

\begin{table*}[!ht]
\begin{center}
\resizebox{\textwidth}{!}{
\begin{tabular}{ccccccccccccccccccccc} \\ \hline
     {Method} & {\rotatebox[origin=l]{70}{road}} 
                          & {\rotatebox[origin=l]{70}{sidewalk}} 
                          & {\rotatebox[origin=l]{70}{building}} 
                          & {\rotatebox[origin=l]{70}{wall}} 
                          & {\rotatebox[origin=l]{70}{fence}} 
                          & {\rotatebox[origin=l]{70}{pole}} 
                          & {\rotatebox[origin=l]{70}{light}} 
                          & {\rotatebox[origin=l]{70}{sign}} 
                          & {\rotatebox[origin=l]{70}{vegetation}} 
                          & {\rotatebox[origin=l]{70}{terrain}} 
                          & {\rotatebox[origin=l]{70}{sky}} 
                          & {\rotatebox[origin=l]{70}{person}} 
                          & {\rotatebox[origin=l]{70}{rider}} 
                          & {\rotatebox[origin=l]{70}{car}} 
                          & {\rotatebox[origin=l]{70}{truck}} 
                          & {\rotatebox[origin=l]{70}{bus}} 
                          & {\rotatebox[origin=l]{70}{train}} 
                          & {\rotatebox[origin=l]{70}{motocycle}} 
                          & {\rotatebox[origin=l]{70}{bicycle}} 
                          & {mIoU} \\  \hline
      \multicolumn{21}{c}{ResNet101 backbone} \\\hline
       AdvEnt\cite{vu_2019_advent} & 89.4 & 33.1 & 81.0 & 26.6 & 26.8 & 27.2 & 33.5 & 24.7 & 83.9 & 36.7 & 78.8 & 58.7 & 30.5 & 84.8 & 38.5 & 44.5 & 1.7 & 31.6 & 32.4 & 45.5 \\
       BDL \cite{Li_19_BDL} & 91.0 & 44.7 & 84.2 & 34.6 & 27.6 & 30.2 & 36.0 & 36.0 & 85.0 & 43.6 & 83.0 & 58.6 & 31.6 & 83.3 & 35.3 & 49.7 & 3.3 & 28.8 & 35.6 & 48.5 \\
       LTIR \cite{Kim_2020_textureInv} & 92.9 & 55.0 & 85.3 & 34.2 & 31.1 & 34.9 & 40.7 & 34.0 & 85.2 & 40.1 & 87.1 & 61.0 & 31.1 & 82.5 & 32.3 & 42.9 & 0.3 & 36.4 & 46.1 & 50.2 \\
       FDA-MBT \cite{yang_2020_FDA} & 92.5 & 53.3 & 82.4 & 26.5 & 27.6 & 36.4 & 40.6 & 38.9 & 82.3 & 39.8 & 78.0 & 62.6 & 34.4 & 84.9 & 34.1 & 53.1 & 16.9 & 27.7 & 46.4 &  50.5 \\
       PCEDA \cite{yang_20_PCEDA} & 91.0& 49.2 & 85.6 & 37.2 & 29.7 & 33.7 & 38.1 & 39.2 & 85.4 & 35.4 & 85.1 & 61.1 & 32.8 & 84.1 & \textbf{45.6} & 46.9 & 0.0 & 34.2 & 44.5 & 50.5 \\
       TPLD \cite{two_phase_ps} & \textbf{94.2} & \textbf{60.5} & 82.8 & 36.6 & 16.6 & 39.3 & 29.0 & 25.5 & 85.6 & 44.9 & 84.4 & 60.6 & 27.4 & 84.1 & 37.0 & 47.0 & \textbf{31.2} & 36.1 & \textbf{50.3} & 51.2 \\ 
       Wang et al. \cite{wang_2021_uncertainty} & 90.5 & 38.7 & \textbf{86.5} & 41.1 & 32.9 & \textbf{40.5} & 48.2 & 42.1 & 86.5 & 36.8 & 84.2 & \textbf{64.5} & \textbf{38.1} & 87.2 & 34.8 & 50.4 & 0.2 & 41.8 & 54.6 & 52.6 \\
       PixMatch \cite{melas_2021_pixmatch} & 91.6 & 51.2 & 84.7 & 37.3 & 29.1 & 24.6 & 31.3 & 37.2 & 86.5 & 44.3 & 85.3 & 62.8 & 22.6 & 87.6 & 38.9 & 52.3 & 0.65 & 37.2 & 50.0 & 50.3 \\
       DPL-Dual (Ensemble) \cite{Cheng_21_dual_path} & 92.8 & 54.4 & 86.2 & 41.6 & 32.7 & 36.4 & \textbf{49.0} & 34.0 & 85.8 & 41.3 & \textbf{86.0} & 63.2 & 34.2 & 87.2 & 39.3 & 44.5 & 18.7 &  \textbf{42.6} & 43.1 & 53.3 \\
       SUDA \cite{SUDA} & 91.1 & 52.3 & 82.9 & 30.1 & 25.7 & 38.0 & 44.9 & 38.2 & 83.9 & 39.1 & 79.2 & 58.4 & 26.4 & 84.5 & 37.7 & 45.6 & 10.1 & 23.1 & 36.0 & 48.8 \\ 
       CaCo \cite{caco} & 91.9 & 54.3 & 82.7 & 31.7 & 25.0 & 38.1 & 46.7 & 39.2 & 82.6 & 39.7 & 76.2 & 63.5 & 23.6 & 85.1 & 38.6 & 47.8 & 10.3 & 23.4 & 35.1 & 49.2 \\ \hline
       Ours & 93.3 & 56.5 & 85.9 & 41.0 & 33.1 & 34.8 & 43.8 &	\textbf{43.8} &	86.6 & 46.5	& 82.5 & 61.1 & 30.4 & 87.0	& 39.7	& 50.7 &	8.8 &	34.9	& 46.8 &	53.0 \\
       Ours (Ensemble) & 93.4 &    55.8 &    86.4 & \textbf{44.4} & \textbf{36.1} & 34.6 & 45.0 & 39.8 &      \textbf{86.9} &   \textbf{48.0} & 84.4 &  61.7 & 30.9 & \textbf{87.7} & 44.9 & \textbf{55.9} & 11.1 & 38.4 &   45.4 & \textbf{54.3} \\ \hline
\end{tabular}}
\end{center}
\caption{Quantitative comparison on GTA5$\rightarrow$Cityscapes. We present per-class IoU and mean IoU (mIoU) obtained using VGG and ResNet101 backbones.}
\label{tab:benchmark_gta2city}
\end{table*}

\begin{table*}[!ht]
\begin{center}
\resizebox{\textwidth}{!}{
\begin{tabular}{ccccccccccccccccccccc} \hline
     {Method} & {\rotatebox[origin=l]{70}{road}} 
                          & {\rotatebox[origin=l]{70}{sidewalk}} 
                          & {\rotatebox[origin=l]{70}{building}} 
                          & {\rotatebox[origin=l]{70}{wall}} 
                          & {\rotatebox[origin=l]{70}{fence}} 
                          & {\rotatebox[origin=l]{70}{pole}} 
                          & {\rotatebox[origin=l]{70}{light}} 
                          & {\rotatebox[origin=l]{70}{sign}} 
                          & {\rotatebox[origin=l]{70}{vegetation}} 
                          & {\rotatebox[origin=l]{70}{sky}} 
                          & {\rotatebox[origin=l]{70}{person}} 
                          & {\rotatebox[origin=l]{70}{rider}} 
                          & {\rotatebox[origin=l]{70}{car}} 
                          & {\rotatebox[origin=l]{70}{bus}} 
                          & {\rotatebox[origin=l]{70}{motocycle}} 
                          & {\rotatebox[origin=l]{70}{bicycle}} 
                          & {mIoU} 
                          & {mIoU*}\\ \hline
       \multicolumn{18}{c}{ResNet101 backbone} \\\hline
       AdvEnt\cite{vu_2019_advent}             & 85.6 & 42.2 & 79.7 & - & - & - & 5.4 & 8.1 & 80.4 & 84.1 & 57.9 & 23.8 & 73.3 & 36.4 & 14.2 & 33.0 & - & 48.0 \\
       LTIR \cite{Kim_2020_textureInv}         & 92.6 & 53.2 & 79.2 & - & - & -  & 1.6 & 7.5 & 78.6 & 84.4 & 52.6 & 20.0 & 82.1 & 34.8 & 14.6 & 39.4 & - & 49.3 \\
       BDL \cite{Li_19_BDL}                    & 86.0 & 46.7 & 80.3 & - & - & - & 14.1 & 11.6 & 79.2 & 81.3 & 54.1 & 27.9 & 73.7 & 42.2 & 25.7 & 45.3 & - & 51.4 \\
       FDA-MBT \cite{yang_2020_FDA}            & 79.3 & 35.0 & 73.2 & - & - & - & 19.9 & 24.0 & 61.7 & 82.6 & 61.4 & \textbf{31.1} & 83.9 & 40.8 & \textbf{38.4} & 51.1 & - & 52.5 \\ 
       PCEDA \cite{yang_20_PCEDA}              & 85.9 & 44.6 & 80.8 & - & - & - & 24.8 & 23.1 & 79.5 & 83.1 & 57.2 & 29.3 & 73.5 & 34.8 &  32.4 & 48.2 & - & 53.6 \\
       TPLD \cite{two_phase_ps}                & 80.9 & 44.3 & 82.2 & 19.9 & 0.3 & \textbf{40.6} & 20.5 & \textbf{30.1} & 77.2 & 80.9 & 60.6 & 25.5 & 84.8 & 41.1 & 24.7 & 43.7 & 47.3 & 53.5 \\
       Wang et al. \cite{wang_2021_uncertainty} & 79.4 & 34.6 & \textbf{83.5} & \textbf{19.3} & \textbf{2.8} & 35.3 & \textbf{32.1} & 26.9 & 78.8 & 79.6 & \textbf{66.6} & 30.3 & 86.1 & 36.6 & 19.5 & \textbf{56.9} & 48.0 & 54.6 \\
       PixMatch \cite{melas_2021_pixmatch} & \textbf{92.5} & \textbf{54.6} & 79.8 & 4.7 & 0.08 & 24.1 & 22.8 & 17.8 & 79.4 & 76.5 & 60.8 & 24.7 & 85.7 & 33.5 & 26.4 & 54.4 & 46.1 & 54.5 \\
       DPL-Dual (Ensemble) \cite{Cheng_21_dual_path}  & 87.5 & 45.7 & 82.8 & 13.3 & 0.6 & 33.2 & 22.0 & 20.1 & 83.1 & 86.0 & 56.6 & 21.9 & 83.1 & 40.3 & 29.8 & 45.7 & 47.0 & 54.2 \\
       SUDA \cite{SUDA} & 83.4 & 36.0 & 71.3 & 8.7 & 0.1 & 26.0 & 18.2 & 26.7 & 72.4 & 80.2 & 58.4 & 30.8 & 80.6 & 38.7 & 36.1 & 46.1 & 44.6 & 52.2 \\
       CaCo \cite{caco} & 87.4 & 48.9 & 79.6 & 8.8 & 0.2 & 30.1 & 17.4 & 28.3 & 79.9 & 81.2 & 56.3 &  24.2 & 78.6 & 39.2 & 28.1 & 48.3 & 46.0 & 53.6 \\ \hline
       Ours	& 85.8 & 41.7	& 82.4	& 7.6	& 1.9	& 33.2	& 26.5	& 18.4	& 83.3	& 86.5	& 62.0	& 29.7	& 83.9	& 52.1	& 34.6	& 51.4	& 48.8	& 56.8 \\ 
       Ours	(Ensemble) & 87.2 & 44.1	& 82.1 & 6.5 & 1.4 & 33.1 & 24.7 & 17.9	& \textbf{83.4} & \textbf{86.6} & 62.4 & 30.4	&  \textbf{86.1}	& \textbf{58.5} & 36.8 & 52.8 & \textbf{49.6}	& \textbf{57.9} \\ \hline
\end{tabular}}
\end{center}
\caption{{Quantitative comparison on SYNTHIA$\rightarrow$Cityscapes.}  We present per-class IoU and mean IoU (mIoU) obtained using VGG and ResNet101 backbones. mIoU and mIoU* are the mean IoU computed on the 16 classes and the 13 subclasses respectively.}
\label{tab:benchmark_syn2city}
\end{table*}
\vspace{-0.3cm}
\section{Conclusions}
In this work we have introduced stochastic translation in the context of UDA and showed that we can reap multiple benefits by acknowledging that certain structures are `lost in translation' across two domains. 
The networks trained directly through stochastic translation clearly outperforms all comparable counterparts, while we have also shown that we retain our edge when combining our approach with more involved UDA approaches such as pseudo-labeling and ensembling. 

\section*{Acknowledgments}
 This research is funded by EPSRC grand EP/N021967/1. We gratefully acknowledge the support of NVIDIA Corporation with the donation of the GPU used for this research.
\bibliography{egbib}

\begin{thebibliography}{44}
\providecommand{\natexlab}[1]{#1}
\providecommand{\url}[1]{\texttt{#1}}
\expandafter\ifx\csname urlstyle\endcsname\relax
  \providecommand{\doi}[1]{doi: #1}\else
  \providecommand{\doi}{doi: \begingroup \urlstyle{rm}\Url}\fi

\bibitem[Almahairi et~al.(2018)Almahairi, Rajeshwar, Sordoni, Bachman, and
  Courville]{Almahairi_18a_augm_cycleGAN}
Amjad Almahairi, Sai Rajeshwar, Alessandro Sordoni, Philip Bachman, and Aaron
  Courville.
\newblock Augmented {C}ycle{GAN}: Learning many-to-many mappings from unpaired
  data.
\newblock In \emph{ICML}, 2018.

\bibitem[Bousmalis et~al.(2017)Bousmalis, Silberman, Dohan, Erhan, and
  Krishnan]{Bousmalis_17_pixel_adapt}
Konstantinos Bousmalis, Nathan Silberman, David Dohan, Dumitru Erhan, and Dilip
  Krishnan.
\newblock Unsupervised pixel-level domain adaptation with generative
  adversarial networks.
\newblock In \emph{CVPR}, 2017.

\bibitem[Chang et~al.(2019)Chang, Wang, Peng, and
  Chiu]{chang_19_all_about_structure}
Wei-Lun Chang, Hui-Po Wang, Wen-Hsiao Peng, and Wei-Chen Chiu.
\newblock All about structure: Adapting structural information across domains
  for boosting semantic segmentation.
\newblock In \emph{CVPR}, 2019.

\bibitem[Chen et~al.(2017)Chen, Papandreou, Kokkinos, Murphy, and
  Yuille]{chen_2017_deeplab}
Liang-Chieh Chen, George Papandreou, Iasonas Kokkinos, Kevin Murphy, and Alan~L
  Yuille.
\newblock {DeepLab}: Semantic image segmentation with deep convolutional nets,
  atrous convolution, and fully connected {CRF}s.
\newblock \emph{TPAMI}, 2017.

\bibitem[Cheng et~al.(2021)Cheng, Wei, Bao, Chen, Wen, and
  Zhang]{Cheng_21_dual_path}
Yiting Cheng, Fangyun Wei, Jianmin Bao, Dong Chen, Fang Wen, and Wenqiang
  Zhang.
\newblock Dual path learning for domain adaptation of semantic segmentation.
\newblock In \emph{ICCV}, 2021.

\bibitem[Chiou et~al.(2020)Chiou, Giganti, Punwani, Kokkinos, and
  Panagiotaki]{Chiou_20_harn_unc}
Eleni Chiou, Francesco Giganti, Shonit Punwani, Iasonas Kokkinos, and
  Eleftheria Panagiotaki.
\newblock Harnessing uncertainty in domain adaptation for mri prostate lesion
  segmentation.
\newblock In \emph{MICCAI}, 2020.

\bibitem[Choi et~al.(2019)Choi, Kim, and Kim]{choi_2019_self_ensembiling}
Jaehoon Choi, Taekyung Kim, and Changick Kim.
\newblock Self-ensembling with gan-based data augmentation for domain
  adaptation in semantic segmentation.
\newblock In \emph{ICCV}, 2019.

\bibitem[Cordts et~al.(2016)Cordts, Omran, Ramos, Rehfeld, Enzweiler, Benenson,
  Franke, Roth, and Schiele]{Cordts_2016_CitySc}
Marius Cordts, Mohamed Omran, Sebastian Ramos, Timo Rehfeld, Markus Enzweiler,
  Rodrigo Benenson, Uwe Franke, Stefan Roth, and Bernt Schiele.
\newblock The cityscapes dataset for semantic urban scene understanding.
\newblock In \emph{CVPR}, 2016.

\bibitem[He et~al.(2016)He, Zhang, Ren, and Sun]{he_2016_resnet}
Kaiming He, Xiangyu Zhang, Shaoqing Ren, and Jian Sun.
\newblock Deep residual learning for image recognition.
\newblock In \emph{CVPR}, 2016.

\bibitem[Hoffman et~al.(2018)Hoffman, Tzeng, Park, Zhu, Isola, Saenko, Efros,
  and Darrell]{Hoffman_18_cycada}
Judy Hoffman, Eric Tzeng, Taesung Park, Jun-Yan Zhu, Phillip Isola, Kate
  Saenko, Alexei Efros, and Trevor Darrell.
\newblock Cycada: Cycle-consistent adversarial domain adaptation.
\newblock In \emph{ICML}, 2018.

\bibitem[Huang et~al.(2022)Huang, Guan, Xiao, Lu, and Shao]{caco}
Jiaxing Huang, Dayan Guan, Aoran Xiao, Shijian Lu, and Ling Shao.
\newblock Category contrast for unsupervised domain adaptation in visual tasks.
\newblock In \emph{CVPR}, 2022.

\bibitem[Huang and Belongie(2017)]{huang2017arbitrary}
Xun Huang and Serge Belongie.
\newblock Arbitrary style transfer in real-time with adaptive instance
  normalization.
\newblock In \emph{ICCV}, 2017.

\bibitem[Huang et~al.(2018)Huang, Liu, Belongie, and Kautz]{Huang_18_munit}
Xun Huang, Ming-Yu Liu, Serge Belongie, and Jan Kautz.
\newblock Multimodal unsupervised image-to-image translation.
\newblock In \emph{ECCV}, 2018.

\bibitem[Kim and Byun(2020)]{Kim_2020_textureInv}
Myeongjin Kim and Hyeran Byun.
\newblock Learning texture invariant representation for domain adaptation of
  semantic segmentation.
\newblock In \emph{CVPR}, 2020.

\bibitem[Lee et~al.(2018)Lee, Tseng, Huang, Singh, and Yang]{Lee_18_DRIT}
Hsin-Ying Lee, Hung-Yu Tseng, Jia-Bin Huang, Maneesh Singh, and Ming-Hsuan
  Yang.
\newblock Diverse image-to-image translation via disentangled representations.
\newblock In \emph{ECCV}, 2018.

\bibitem[Li et~al.(2019)Li, Yuan, and Vasconcelos]{Li_19_BDL}
Yunsheng Li, Lu~Yuan, and Nuno Vasconcelos.
\newblock Bidirectional learning for domain adaptation of semantic
  segmentation.
\newblock In \emph{CVPR}, 2019.

\bibitem[Long et~al.(2015)Long, Shelhamer, and Darrell]{long_2015_fully}
Jonathan Long, Evan Shelhamer, and Trevor Darrell.
\newblock Fully convolutional networks for semantic segmentation.
\newblock In \emph{CVPR}, 2015.

\bibitem[Luo et~al.(2019)Luo, Zheng, Guan, Yu, and
  Yang]{luo_2019_taking_a_closer}
Yawei Luo, Liang Zheng, Tao Guan, Junqing Yu, and Yi~Yang.
\newblock Taking a closer look at domain shift: Category-level adversaries for
  semantics consistent domain adaptation.
\newblock In \emph{CVPR}, 2019.

\bibitem[Ma et~al.(2021)Ma, Lin, Wu, and Yu]{ma_2021_coarse_to_fine}
Haoyu Ma, Xiangru Lin, Zifeng Wu, and Yizhou Yu.
\newblock Coarse-to-fine domain adaptive semantic segmentation with photometric
  alignment and category-center regularization.
\newblock In \emph{CVPR}, 2021.

\bibitem[Mei et~al.(2020)Mei, Zhu, Zou, and Zhang]{mei_2020_instance}
Ke~Mei, Chuang Zhu, Jiaqi Zou, and Shanghang Zhang.
\newblock Instance adaptive self-training for unsupervised domain adaptation.
\newblock In \emph{CVPR}, 2020.

\bibitem[Melas-Kyriazi and Manrai(2021)]{melas_2021_pixmatch}
Luke Melas-Kyriazi and Arjun~K Manrai.
\newblock Pixmatch: Unsupervised domain adaptation via pixelwise consistency
  training.
\newblock In \emph{CVPR}, 2021.

\bibitem[Pan et~al.(2020)Pan, Shin, Rameau, Lee, and
  Kweon]{pan_2020_intra_domain}
Fei Pan, Inkyu Shin, Francois Rameau, Seokju Lee, and In~So Kweon.
\newblock Unsupervised intra-domain adaptation for semantic segmentation
  through self-supervision.
\newblock In \emph{CVPR}, 2020.

\bibitem[Richter et~al.(2016)Richter, Vineet, Roth, and
  Koltun]{Richter_2016_GTA5}
Stephan~R. Richter, Vibhav Vineet, Stefan Roth, and Vladlen Koltun.
\newblock Playing for data: {G}round truth from computer games.
\newblock In \emph{ECCV}, 2016.

\bibitem[Ros et~al.(2016)Ros, Sellart, Materzynska, Vazquez, and
  Lopez]{Ros_2016_SYNTHIA}
German Ros, Laura Sellart, Joanna Materzynska, David Vazquez, and Antonio~M.
  Lopez.
\newblock The synthia dataset: A large collection of synthetic images for
  semantic segmentation of urban scenes.
\newblock In \emph{CVPR}, June 2016.

\bibitem[Saito et~al.(2018)Saito, Watanabe, Ushiku, and
  Harada]{saito_2018_maximum_classifier}
Kuniaki Saito, Kohei Watanabe, Yoshitaka Ushiku, and Tatsuya Harada.
\newblock Maximum classifier discrepancy for unsupervised domain adaptation.
\newblock In \emph{CVPR}, 2018.

\bibitem[Shin et~al.(2020)Shin, Woo, Pan, and Kweon]{two_phase_ps}
Inkyu Shin, Sanghyun Woo, Fei Pan, and In~So Kweon.
\newblock Two-phase pseudo label densification for self-training based domain
  adaptation.
\newblock In \emph{ECCV}, 2020.

\bibitem[Simonyan and Zisserman(2014)]{simonyan_2014_VGG}
Karen Simonyan and Andrew Zisserman.
\newblock Very deep convolutional networks for large-scale image recognition.
\newblock In \emph{CoRR, abs/1409.1556}, 2014.

\bibitem[Tsai et~al.(2018)Tsai, Hung, Schulter, Sohn, Yang, and
  Chandraker]{tsai_2018_AdaptSegNet}
Yi-Hsuan Tsai, Wei-Chih Hung, Samuel Schulter, Kihyuk Sohn, Ming-Hsuan Yang,
  and Manmohan Chandraker.
\newblock Learning to adapt structured output space for semantic segmentation.
\newblock In \emph{CVPR}, 2018.

\bibitem[Tsai et~al.(2019)Tsai, Sohn, Schulter, and
  Chandraker]{tsai_2019_discriminative_patch}
Yi-Hsuan Tsai, Kihyuk Sohn, Samuel Schulter, and Manmohan Chandraker.
\newblock Domain adaptation for structured output via discriminative patch
  representations.
\newblock In \emph{CVPR}, 2019.

\bibitem[Vu et~al.(2019)Vu, Jain, Bucher, Cord, and P{\'e}rez]{vu_2019_advent}
Tuan-Hung Vu, Himalaya Jain, Maxime Bucher, Matthieu Cord, and Patrick
  P{\'e}rez.
\newblock {ADVENT}: Adversarial entropy minimization for domain adaptation in
  semantic segmentation.
\newblock In \emph{CVPR}, 2019.

\bibitem[Wang et~al.(2020{\natexlab{a}})Wang, Shen, Zhang, Duan, and
  Mei]{wang_2020_classes_matter}
Haoran Wang, Tong Shen, Wei Zhang, Ling-Yu Duan, and Tao Mei.
\newblock Classes matter: A fine-grained adversarial approach to cross-domain
  semantic segmentation.
\newblock In \emph{ECCV}, 2020{\natexlab{a}}.

\bibitem[Wang et~al.(2021)Wang, Peng, and Zhang]{wang_2021_uncertainty}
Yuxi Wang, Junran Peng, and ZhaoXiang Zhang.
\newblock Uncertainty-aware pseudo label refinery for domain adaptive semantic
  segmentation.
\newblock In \emph{ICCV}, 2021.

\bibitem[Wang et~al.(2020{\natexlab{b}})Wang, Yu, Wei, Feris, Xiong, Hwu,
  Huang, and Shi]{Wang_2020_diff_treat}
Zhonghao Wang, Mo~Yu, Yunchao Wei, Rogerio Feris, Jinjun Xiong, Wen-mei Hwu,
  Thomas~S. Huang, and Honghui Shi.
\newblock Differential treatment for stuff and things: A simple unsupervised
  domain adaptation method for semantic segmentation.
\newblock In \emph{CVPR}, 2020{\natexlab{b}}.

\bibitem[Wu et~al.(2018)Wu, Han, Lin, Uzunbas, Goldstein, Lim, and
  Davis]{wu2018dcan}
Zuxuan Wu, Xintong Han, Yen-Liang Lin, Mustafa~Gokhan Uzunbas, Tom Goldstein,
  Ser~Nam Lim, and Larry~S Davis.
\newblock Dcan: Dual channel-wise alignment networks for unsupervised scene
  adaptation.
\newblock In \emph{ECCV}, 2018.

\bibitem[Yang et~al.(2020{\natexlab{a}})Yang, An, Wang, Zhu, Yan, and
  Huang]{yang_2020_label_driv_rec}
Jinyu Yang, Weizhi An, Sheng Wang, Xinliang Zhu, Chaochao Yan, and Junzhou
  Huang.
\newblock Label-driven reconstruction for domain adaptation in semantic
  segmentation.
\newblock In \emph{ECCV}, 2020{\natexlab{a}}.

\bibitem[Yang and Soatto(2020)]{yang_2020_FDA}
Yanchao Yang and Stefano Soatto.
\newblock {FDA}: Fourier domain adaptation for semantic segmentation.
\newblock In \emph{CVPR}, 2020.

\bibitem[Yang et~al.(2020{\natexlab{b}})Yang, Lao, Sundaramoorthi, and
  Soatto]{yang_20_PCEDA}
Yanchao Yang, Dong Lao, Ganesh Sundaramoorthi, and Stefano Soatto.
\newblock Phase consistent ecological domain adaptation.
\newblock In \emph{CVPR}, 2020{\natexlab{b}}.

\bibitem[Zhang et~al.(2022)Zhang, Huang, Tian, and Lu]{SUDA}
Jingyi Zhang, Jiaxing Huang, Zichen Tian, and Shijian Lu.
\newblock Spectral unsupervised domain adaptation for visual recognition.
\newblock In \emph{CVPR}, 2022.

\bibitem[Zhang et~al.(2021)Zhang, Zhang, Zhang, Chen, Wang, and
  Wen]{zhang_2021_proda}
Pan Zhang, Bo~Zhang, Ting Zhang, Dong Chen, Yong Wang, and Fang Wen.
\newblock Prototypical pseudo label denoising and target structure learning for
  domain adaptive semantic segmentation.
\newblock In \emph{CVPR}, 2021.

\bibitem[Zhang et~al.(2019)Zhang, Zhang, Liu, and
  Tao]{zhang_2019_category_anchor}
Qiming Zhang, Jing Zhang, Wei Liu, and Dacheng Tao.
\newblock Category anchor-guided unsupervised domain adaptation for semantic
  segmentation.
\newblock \emph{NeurIPS}, 2019.

\bibitem[Zhu et~al.(2017{\natexlab{a}})Zhu, Park, Isola, and
  Efros]{Zhu_2017_CycleGAN}
Jun-Yan Zhu, Taesung Park, Phillip Isola, and Alexei~A Efros.
\newblock Unpaired image-to-image translation using cycle-consistent
  adversarial networks.
\newblock In \emph{ICCV}, 2017{\natexlab{a}}.

\bibitem[Zhu et~al.(2017{\natexlab{b}})Zhu, Zhang, Pathak, Darrell, Efros,
  Wang, and Shechtman]{Zhu_17_BicycleGAN}
Jun-Yan Zhu, Richard Zhang, Deepak Pathak, Trevor Darrell, Alexei~A Efros,
  Oliver Wang, and Eli Shechtman.
\newblock Toward multimodal image-to-image translation.
\newblock In \emph{NIPS}, 2017{\natexlab{b}}.

\bibitem[Zou et~al.(2018)Zou, Yu, Kumar, and
  Wang]{zou_2018_class_balanced_self_training}
Yang Zou, Zhiding Yu, BVK Kumar, and Jinsong Wang.
\newblock Unsupervised domain adaptation for semantic segmentation via
  class-balanced self-training.
\newblock In \emph{ECCV}, 2018.

\bibitem[Zou et~al.(2019)Zou, Yu, Liu, Kumar, and
  Wang]{zou_2019_confidence_regularized}
Yang Zou, Zhiding Yu, Xiaofeng Liu, BVK Kumar, and Jinsong Wang.
\newblock Confidence regularized self-training.
\newblock In \emph{ICCV}, 2019.

\end{thebibliography}

\section*{Supplementary}

\subsection*{Overview}
We provide additional details about the training procedure and additional results including quantitative results and qualitative results.

\subsection*{Training procedure}
In \refsec{sec:st_tr} we provide some additional details regarding the training procedure of the stochastic translation network while in \refsec{sec:pseudo} we provide additional details about the pseudo-labeling. Finally, in \refsec{sec:train_process} we summarize in more detail the entire training process. 
\subsubsection*{Stochastic translation for unsupervised domain adaptation}
\label{sec:st_tr}
\begin{figure}[!t]
\centering
\includegraphics[width=1\textwidth]{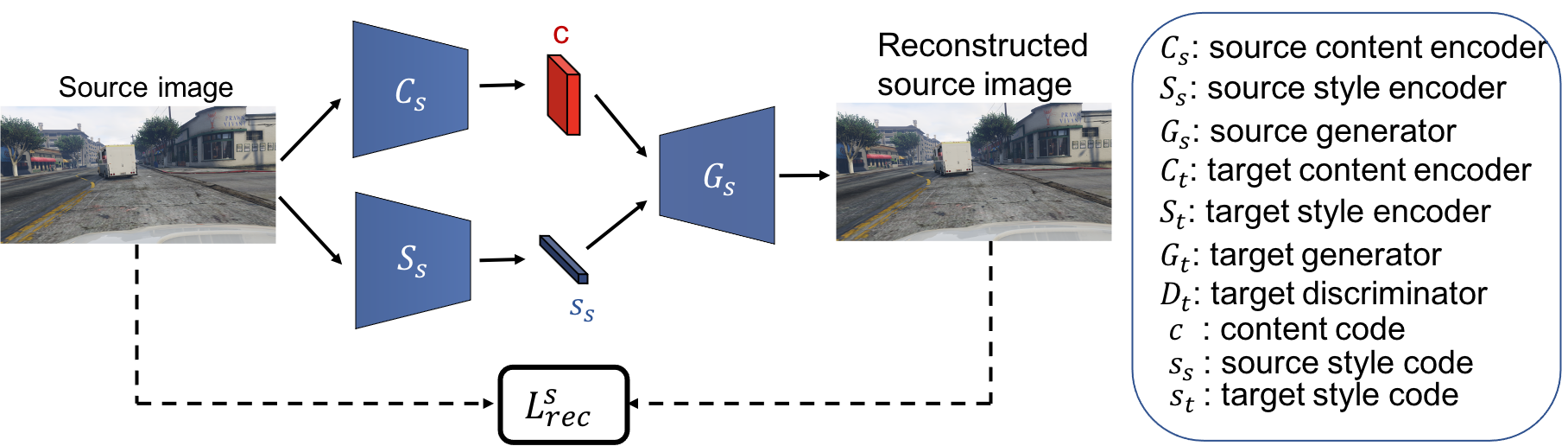}
\hfill
\includegraphics[width=1\textwidth]{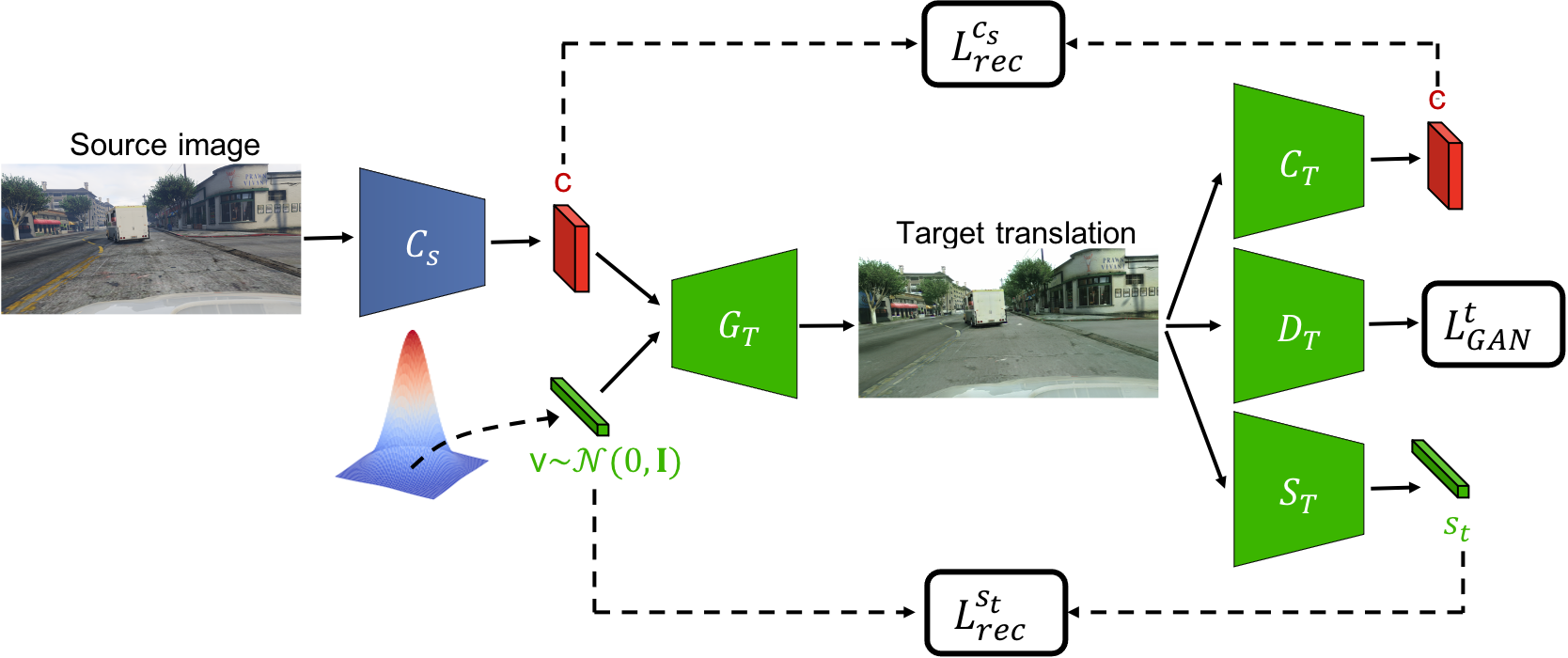}
\caption{Stochastic translation~\cite{Huang_18_munit} from the source to the target domain: with-in domain reconstruction (top) and cross-domain translation (bottom) allows us to reconstruct the input and pass the content from the source image to its counterparts respectively. The target cycle is omitted for clarity.}
\label{fig:st_transl}
\end{figure}

 The stochastic translation network is based on MUNIT~\cite{Huang_18_munit}. We have described it in the main paper but we also describe it here in more detail. For a more extensive presentation we refer the reader to \cite{Huang_18_munit}. As it is illustrated in \reffig{fig:st_transl}, the stochastic translation network consists of content encoders $\{C_s, C_t\}$, style encoders $\{S_s, S_t\}$, generators $\{G_s, G_t\}$ and domain discriminators $\{D_s, D_t\}$ for the source domain $s$, and the target domain $t$ respectively.

Given a source domain image $~x \in \mathcal{S}$, we start by extracting a domain-invariant content code $c=C_s(x)$ and a domain-specific style code $s_s=S_s(x)$. Then, we perform with-in domain reconstruction (\reffig{fig:st_transl}, top) and cross-domain translation (\reffig{fig:st_transl}, bottom). We reconstruct the original image $x$, using the source generator $G_s$ that takes as input at the first layer the content code $c$ and its subsequent layers are modulated by Adaptive Instance Normalization (AdaIN)~\cite{huang2017arbitrary} driven by the style code $s_s$. This amounts in minimizing the following objective function:
\begin{equation*}
L_{rec}^s = \sum_{x\in \mathcal{S}} \| x - G_{s}(\content_s(x),\style_s(x))\|.
\end{equation*}

We perform translation from the source domain to the target domain by passing the content code $c$ as input to the target generator $G_t$, whose subsequent layers are modulated by AdaIn driven by a random style code $~ \noise \sim \normal$. This results in the following stochastic translation function from the source domain to the target domain:
\begin{equation*}
\transd{x}{\noise} \doteq G_{t}(\content_s(x),\noise).
\end{equation*}

We ensure that the resulting translation matches the distribution of the target domain data by employing the following GAN objective: 
\begin{equation*}
L_{GAN}^t = \sum_{x\in \mathcal{T}} \log D_t(x) + \sum_{x\in \mathcal{S}} \log(1-D_t(\transd{x}{\noise})),
\end{equation*}
where $\mathcal{T}$, $\mathcal{S}$ the target and source datasets respectively and $D_t$ the adversarial discriminator for the the target domain $t$.

We ensure that the content codes $c$ of the source image and the translated image are aligned by minimize the following objective:
\begin{equation*}
L_{rec}^{c_s} = \sum_{x\in \mathcal{S}} \|\content_t(G_{t}(\content_s(x),\noise)) - \content_s(x)\|_2.
\end{equation*}

Similarly, to align the target style code with the Gaussian prior distribution we use an objective of the following form:
\begin{equation*}
L_{rec}^{s_t} = \sum_{x\in \mathcal{S}} \|\style_t(G_{t}(\content_s(x),\noise)) - \noise\|_2.
\end{equation*}
As we described in detail in the main paper we ensure that the semantics are preserved during translation  using the following objective function: 
\begin{equation*}
L_{sem}^s = \loss_{ce}(F(\transd{x}{\noise}), p),
\label{eq:semloss_1}
\end{equation*}
where $\loss_{ce}$, the cross-entropy loss, $F$ a segmentation network trained on both source and target data, $F(\transd{x}{\noise})$ the softmax output given the translated image $\transd{x}{\noise}$ and $p = \textrm{argmax}(F(x))$ the predicted labels for the source image $~x \in \mathcal{S}$.

The exact same procedure is followed for translating from the target  to the source domain and the corresponding loss terms $L_{rec}^t$, $L_{GAN}^s$, $L_{rec}^{c_t}$, $L_{rec}^{s_s}$, $L_{sem}^t$ are defined similarly. 

The full objective is given by 
\begin{equation}
\bal
\min_{\substack{C_s,  S_s, G_s \\ C_t, S_t, G_t}} \max_{ D_s, D_t} 
& \quad \lambda_{x} ( L_{rec}^{s} +  L_{rec}^{t}) +  \lambda_{GAN} (L_{GAN}^s +  L_{GAN}^t)  \\
& \quad + \lambda_{c} ( L_{rec}^{c_s} + L_{rec}^{c_t}) +\lambda_{s} ( L_{rec}^{s_s} +  L_{rec}^{s_t})  \\
& \quad + \lambda_{sem} ( L_{sem}^{s} +  L_{sem}^{t}),
\eal
\end{equation}
where $\lambda_{GAN}$, $\lambda_{x}$, $\lambda_{c}$, $\lambda_{s}$, $\lambda_{sem}$ are weights that control the importance of each term.  In our experiments, we set the weights as follows: $\lambda_{GAN}=1$, $\lambda_{x}=10$, $\lambda_{c}=1$, $\lambda_{s}=1$, $\lambda_{sem}=1$.

\subsubsection*{Robust pseudo-label generation} 
\label{sec:pseudo}
As we mentioned in Sec. 3.4 of the main paper, we rely on three complementary networks to generate robust pseudo-labels. We train two target networks $F_{t,\sigma^2=1}$, $F_{t,\sigma^2=10}$; one with the variance left intact and the other with the variance scaled by 10. We also train a source network, $F_s$, and exploit multiple samples to obtain a better estimate of the pseudo-labels as we described in Sec. 3.3 of the main paper.

The enhanced probability map used to generate pseudo-labels is obtained by the weighted average of the predictions of the three networks: 
\begin{equation}
\hat{y} = \frac{1}{3} \hat{y_s} + \frac{1}{3}\hat{y}_{t,\sigma^2=1} + \frac{1}{3}\hat{y}_{t,\sigma^2=10},
\end{equation}
where $\hat{y_s}$, $\hat{y}_{t,\sigma^2=1}$, $\hat{y}_{t,\sigma^2=10}$ indicates the pixel-level posterior distribution on labels obtained by $F_s$, $F_{t,\sigma^2=1}$, $F_{t,\sigma^2=10}$ respectively. 

We assign pseudo-labels to samples for which the dominant class has a score above a certain threshold $\theta$. Similarly to~\cite{zou_2018_class_balanced_self_training} we use class-wise confidence thresholds to assign pseudo-labels. In particular for each class $c$, the threshold $\theta_c$ equals the probability ranked at $r * N_c$, where $N_c$ is the number of pixels predicted to belong in class $c$ and $r$ is the proportion of pseudo-labels we want to retain. In cases where $\theta_c > 0.9$, we set $\theta_c = 0.9$. In \refsec{sec:add_results_quant_res} we provide results for the selection of $r$ based on the mean intersection-over-union (mIoU) of the validation set.
\newcommand{\normalnew}{\mathcal{N}(\mathbf{0},10\mathbf{I})}

\subsubsection*{Training process}
\label{sec:train_process}
Algorithm \ref{algr:train_proc} summarizes the training process. Initially we train a segmentation network, $F$, operating on both domains and use this to impose a semantic consistency constrain to the stochastic translation network trained using Eq. 11. Using translated images obtained by the stochastic translation network, we train two target-domain networks and one source-domain network. We generate robust pseudo-labels for the target-domain data by combining the predictions of the three models. The pseudo-labels are used in the next round of training as supervision for the networks when they are driven by target-domain data. We perform two rounds (R) of pseudo-labeling and training.
\renewcommand{\algorithmicrequire}{\textbf{Input:}}
\renewcommand{\algorithmicensure}{\textbf{Output:}}
\begin{algorithm}[ht]
	\small
\caption{Training process}\label{algr:train_proc}
\begin{algorithmic}
    \Require ${\mathcal{S}}$, ${\mathcal{T}}$
\Ensure{${F_{t,\sigma^2=1}^{(R=2)}}$, ${F_{t,\sigma^2=10}^{(R=2)}}$, ${F_{
s}^{(R=2)}}$}
\State {train $F$ with Eq. 2 (main paper)}
\State {train the stochastic translation network with Eq. 11}
\State {train $F_{t,\sigma^2=1}^{(R=0)}$ with Eq. 4 (main paper), where $\noise \sim \normal$} 
\State {train $F_{t,\sigma^2=10}^{(R=0)}$ with Eq. 4 (main paper), where $\noise \sim \normalnew$} 
\State {train $F_{s}^{(R=0)}$ with Eq. 9 (main paper), where $\noise \sim \normal$ }
\For{$i \gets 1$ to $2$} \\
     \hspace{\algorithmicindent} generate $\hat{y}$ with Eq. 12 \\
     \hspace{\algorithmicindent}train $F_{t,\sigma^2=1}^{(R=i)}$ with Eq. 8 (main paper), where $\noise \sim \normal$ \\
    \hspace{\algorithmicindent}train $F_{t,\sigma^2=10}^{(R=i)}$ with Eq. 8 (main paper), where $\noise \sim \normalnew$ \\
   \hspace{\algorithmicindent}train ${F_{\mathcal{S}}^{(R=i)}}$ with source-domain counterpart of Eq. 8 (main paper), where $\noise \sim \normal$
\EndFor
\end{algorithmic}
\end{algorithm}

\subsection*{Additional results}
\label{sec:add_results}
We provide results (mIoU and per-class IoU) obtained using both DeepLab-V2 with ResNet-101 and FCN-8s with VGG-16 and compare with state-of-the-art methods on both GTA-to-Cityscapes and SYNTHIA-to-Cityscapes benchmarks. mIoU results for DeepLab-V2 with ResNet-101 for both GTA-to-Cityscapes and SYNTHIA-to-Cityscapes benchmarks have already been presented in the main paper.

We also report results from additional ablation studies and class-wise IoU for ablation studies already presented in the main paper. We report results on GTA-to-Cityscapes using
DeepLab-V2 with ResNet-101.

\subsubsection*{Quantitative results}
\label{sec:add_results_quant_res}
\noindent \textbf{Comparison with state-of-the-art methods on GTA-to-Cityscapes and SYNTHIA-to-Cityscapes benchmarks.}\\
In Table~\ref{tab:benchmark_gta2city_sup} we provide results (per-class IoU and mIoU) and comparison with recent state-of-the-art methods for the GTA-to-Cityscapes benchmark. (mIoU results are also reported in Table 4 of the main paper). Our
results show that our method achieves state-of-the-art performance and outperforms previous methods.
\begin{table*}[!ht]
\begin{center}
\resizebox{\textwidth}{!}{
\begin{tabular}{ccccccccccccccccccccc} \\ \hline
     {Method} & {\rotatebox[origin=l]{70}{road}} 
                          & {\rotatebox[origin=l]{70}{sidewalk}} 
                          & {\rotatebox[origin=l]{70}{building}} 
                          & {\rotatebox[origin=l]{70}{wall}} 
                          & {\rotatebox[origin=l]{70}{fence}} 
                          & {\rotatebox[origin=l]{70}{pole}} 
                          & {\rotatebox[origin=l]{70}{light}} 
                          & {\rotatebox[origin=l]{70}{sign}} 
                          & {\rotatebox[origin=l]{70}{vegetation}} 
                          & {\rotatebox[origin=l]{70}{terrain}} 
                          & {\rotatebox[origin=l]{70}{sky}} 
                          & {\rotatebox[origin=l]{70}{person}} 
                          & {\rotatebox[origin=l]{70}{rider}} 
                          & {\rotatebox[origin=l]{70}{car}} 
                          & {\rotatebox[origin=l]{70}{truck}} 
                          & {\rotatebox[origin=l]{70}{bus}} 
                          & {\rotatebox[origin=l]{70}{train}} 
                          & {\rotatebox[origin=l]{70}{motocycle}} 
                          & {\rotatebox[origin=l]{70}{bicycle}} 
                          & {mIoU} \\  \hline
        \multicolumn{21}{c}{VGG backbone} \\\hline
        AdaptSegNet\cite{tsai_2018_AdaptSegNet} & 87.3 & 29.8 & 78.6 & 21.1 & 18.2 & 22.5 & 21.5 & 11.0 & 79.7 & 29.6 & 71.3 & 46.8 & 6.5 & 80.1 & 23.0 & 26.9 & 0.0 & 10.6 & 0.3 & 35.0 \\
       AdvEnt\cite{tsai_2018_AdaptSegNet} & 86.9 & 28.7 & 78.7 & 28.5 & 25.2 & 17.1 & 20.3 & 10.9 & 80.0 & 26.4 & 70.2 & 47.1 & 8.4 & 81.5 & 26.0 & 17.2 & 18.9 & 11.7 & 1.6 & 36.1 \\
       BDL \cite{Li_19_BDL} & 89.2 & 40.9 & 81.2 & 29.1 & 19.2 & 14.2 & 29.0 & 19.6 & 83.7 & 35.9 & 80.7 & 54.7 & 23.3 & 82.7 & 25.8 & 28.0 & 2.3 & 25.7 & 19.9 & 41.3 \\
       LTIR \cite{Kim_2020_textureInv} & \textbf{92.5} & \textbf{54.5} & 83.9 & 34.5 & 25.5 & 31.0 & 30.4 & 18.0 & 84.1 & 39.6 & \textbf{83.9} & 53.6 & 19.3 & 81.7 & 21.1 & 13.6 & 17.7 & 12.3 & 6.5 & 42.3 \\
       FDA-MBT \cite{yang_2020_FDA} & 86.1 & 35.1 & 80.6 & 30.8 & 20.4 & 27.5 & 30.0 & 26.0 & 82.1 & 30.3 & 73.6 & 52.5 & 21.7 & 81.7 & 24.0 & 30.5 & \textbf{29.9} & 14.6 & 24.0 &  42.2 \\
       PCEDA \cite{yang_20_PCEDA} & 90.7 & 49.8 & 81.9 & 23.4 & 18.5 & \textbf{37.3} & 35.5 & 34.3 & 82.9 & 36.5 & 75.8 & \textbf{61.8} & 12.4 & 83.2 & 19.2 & 26.1 & 4.0 & 14.3 &  21.8 &  42.6 \\
       DPL-Dual (Ensemble) \cite{Cheng_21_dual_path}  & 89.2 & 44.0 & 83.5 & \textbf{35.0} & 24.7 & 27.8 & \textbf{38.3} & 25.3 & 84.2 & 39.5 & 81.6 & 54.7 & \textbf{25.8} & 83.3 & 29.3 & 49.0 & 5.2 & 30.2 & 32.6 & 46.5 \\ \hline
       Ours &  91.1 & 43.2 & 84.1 & 	34.6 &	25.5 &	25.8 &	33.7 &	31.3 &	84.7 &	\textbf{44.9} &	83.1 & 55.3 &  23.5	& 81.6 &	23.1 &	34.3 & 6.3 & \textbf{32.7} & 34.8 & 46.0 \\
       Ours (Ensemble)  & 91.0 & 40.7 & \textbf{84.7} & 33.8	& \textbf{27.1} &	30.9 &	33.1 &	\textbf{35.1} &	\textbf{85.3} &	44.7 & 82.9 &	56.8 & 	23.4 & 	\textbf{86.2} & \textbf{36.5} & \textbf{50.3} & 2.8 & 27.8 & \textbf{36.6} & 	\textbf{47.9} \\ \hline
       
      \multicolumn{21}{c}{ResNet101 backbone} \\\hline
       AdvEnt\cite{vu_2019_advent} & 89.4 & 33.1 & 81.0 & 26.6 & 26.8 & 27.2 & 33.5 & 24.7 & 83.9 & 36.7 & 78.8 & 58.7 & 30.5 & 84.8 & 38.5 & 44.5 & 1.7 & 31.6 & 32.4 & 45.5 \\
       BDL \cite{Li_19_BDL} & 91.0 & 44.7 & 84.2 & 34.6 & 27.6 & 30.2 & 36.0 & 36.0 & 85.0 & 43.6 & 83.0 & 58.6 & 31.6 & 83.3 & 35.3 & 49.7 & 3.3 & 28.8 & 35.6 & 48.5 \\
       LTIR \cite{Kim_2020_textureInv} & 92.9 & 55.0 & 85.3 & 34.2 & 31.1 & 34.9 & 40.7 & 34.0 & 85.2 & 40.1 & 87.1 & 61.0 & 31.1 & 82.5 & 32.3 & 42.9 & 0.3 & 36.4 & 46.1 & 50.2 \\
       FDA-MBT \cite{yang_2020_FDA} & 92.5 & 53.3 & 82.4 & 26.5 & 27.6 & 36.4 & 40.6 & 38.9 & 82.3 & 39.8 & 78.0 & 62.6 & 34.4 & 84.9 & 34.1 & 53.1 & 16.9 & 27.7 & 46.4 &  50.5 \\
       PCEDA \cite{yang_20_PCEDA} & 91.0& 49.2 & 85.6 & 37.2 & 29.7 & 33.7 & 38.1 & 39.2 & 85.4 & 35.4 & 85.1 & 61.1 & 32.8 & 84.1 & \textbf{45.6} & 46.9 & 0.0 & 34.2 & 44.5 & 50.5 \\
       TPLD \cite{two_phase_ps} & \textbf{94.2} & \textbf{60.5} & 82.8 & 36.6 & 16.6 & 39.3 & 29.0 & 25.5 & 85.6 & 44.9 & 84.4 & 60.6 & 27.4 & 84.1 & 37.0 & 47.0 & \textbf{31.2} & 36.1 & \textbf{50.3} & 51.2 \\ 
       Wang et al. \cite{wang_2021_uncertainty} & 90.5 & 38.7 & \textbf{86.5} & 41.1 & 32.9 & \textbf{40.5} & 48.2 & 42.1 & 86.5 & 36.8 & 84.2 & \textbf{64.5} & \textbf{38.1} & 87.2 & 34.8 & 50.4 & 0.2 & 41.8 & 54.6 & 52.6 \\
       PixMatch \cite{melas_2021_pixmatch} & 91.6 & 51.2 & 84.7 & 37.3 & 29.1 & 24.6 & 31.3 & 37.2 & 86.5 & 44.3 & 85.3 & 62.8 & 22.6 & 87.6 & 38.9 & 52.3 & 0.65 & 37.2 & 50.0 & 50.3 \\
       DPL-Dual (Ensemble) \cite{Cheng_21_dual_path} & 92.8 & 54.4 & 86.2 & 41.6 & 32.7 & 36.4 & \textbf{49.0} & 34.0 & 85.8 & 41.3 & \textbf{86.0} & 63.2 & 34.2 & 87.2 & 39.3 & 44.5 & 18.7 &  \textbf{42.6} & 43.1 & 53.3 \\ 
        SUDA \cite{SUDA} & 91.1 & 52.3 & 82.9 & 30.1 & 25.7 & 38.0 & 44.9 & 38.2 & 83.9 & 39.1 & 79.2 & 58.4 & 26.4 & 84.5 & 37.7 & 45.6 & 10.1 & 23.1 & 36.0 & 48.8 \\ 
      CaCo \cite{caco} & 91.9 & 54.3 & 82.7 & 31.7 & 25.0 & 38.1 & 46.7 & 39.2 & 82.6 & 39.7 & 76.2 & 63.5 & 23.6 & 85.1 & 38.6 & 47.8 & 10.3 & 23.4 & 35.1 & 49.2 \\ \hline
       Ours & 93.3 & 56.5 & 85.9 & 41.0 & 33.1 & 34.8 & 43.8 &	\textbf{43.8} &	86.6 & 46.5	& 82.5 & 61.1 & 30.4 & 87.0	& 39.7	& 50.7 &	8.8 &	34.9	& 46.8 &	53.0 \\
       Ours (Ensemble) & 93.4 &    55.8 &    86.4 & \textbf{44.4} & \textbf{36.1} & 34.6 & 45.0 & 39.8 &      \textbf{86.9} &   \textbf{48.0} & 84.4 &  61.7 & 30.9 & \textbf{87.7} & 44.9 & \textbf{55.9} & 11.1 & 38.4 &   45.4 & \textbf{54.3} \\ \hline
\end{tabular}}
\end{center}
\caption{Quantitative comparison on GTA5$\rightarrow$Cityscapes. We present per-class IoU and mean IoU (mIoU) obtained using VGG and ResNet101 backbones.}
\label{tab:benchmark_gta2city_sup}
\end{table*}

In Table~\ref{tab:benchmark_syn2city_sup} we provide results (per-class IoU and mIoU) and comparison with recent state-of-the-art methods for the SYNTHIA-to-Cityscapes benchmark. (mIoU results are also reported in Table 5 of the main paper). Our
results show that our method achieves state-of-the-art performance and outperforms previous methods.
\begin{table*}[!ht]
\begin{center}
\resizebox{\textwidth}{!}{
\begin{tabular}{ccccccccccccccccccccc} \hline
     {Method} & {\rotatebox[origin=l]{70}{road}} 
                          & {\rotatebox[origin=l]{70}{sidewalk}} 
                          & {\rotatebox[origin=l]{70}{building}} 
                          & {\rotatebox[origin=l]{70}{wall}} 
                          & {\rotatebox[origin=l]{70}{fence}} 
                          & {\rotatebox[origin=l]{70}{pole}} 
                          & {\rotatebox[origin=l]{70}{light}} 
                          & {\rotatebox[origin=l]{70}{sign}} 
                          & {\rotatebox[origin=l]{70}{vegetation}} 
                          & {\rotatebox[origin=l]{70}{sky}} 
                          & {\rotatebox[origin=l]{70}{person}} 
                          & {\rotatebox[origin=l]{70}{rider}} 
                          & {\rotatebox[origin=l]{70}{car}} 
                          & {\rotatebox[origin=l]{70}{bus}} 
                          & {\rotatebox[origin=l]{70}{motocycle}} 
                          & {\rotatebox[origin=l]{70}{bicycle}} 
                          & {mIoU} 
                          & {mIoU*}\\ \hline
     \multicolumn{18}{c}{VGG backbone} \\\hline
       AdvEnt\cite{vu_2019_advent} & 67.9 & 29.4 & 71.9 & 6.3 & 0.3 & 19.9 & 0.6 & 2.6 & 74.9 & 74.9 & 35.4 & 9.6 & 67.8 & 21.4 & 4.1 & 15.5 & 31.4 & 36.6 \\
       BDL \cite{Li_19_BDL} & 72.0 & 30.3 & 74.5 & 0.1 & 0.3 & 24.6 & 10.2 & 25.2 & 80.5 & 80.0 & 54.7 & 23.2 & 72.7 & 24.0 & 7.5 & 44.9 & 39.0 & 46.1 \\
       FDA-MBT \cite{yang_2020_FDA} & 84.2 & 35.1 & 78.0 & \textbf{6.1} & 0.4 & 27.0 & 8.5 & 22.1 & 77.2 & 79.6 & 55.5 & 19.9 & 74.8 & 24.9 & 14.3 & 40.7 & 40.5 & 47.3 \\ 
       PCEDA \cite{yang_20_PCEDA} & 79.7 & 35.2 & 78.7 & 1.4 & 0.6 & 23.1 & 10.0 & 28.9 & 79.6 & 81.2 & 51.2 & \textbf{25.1} & 72.2 & 24.1 & 16.7 & \textbf{50.4} & 41.1 & 48.7 \\
       DPL-Dual (Ensemble) \cite{Cheng_21_dual_path} & 83.5 & 38.2 & \textbf{80.4} & 1.3 & \textbf{1.1} & \textbf{29.1} & \textbf{20.2} & 32.7 & 81.8 & 83.6 & 55.9 & 20.3 & 79.4 & 26.6 & 7.4 & 46.2 & 43.0 & 50.5 \\ \hline
       Ours & 83.3 &  40.9 &  80.3 & 1.4 & 0.6 &  24.8 &  16.9 &  31.1 &  \textbf{82.4} &  \textbf{84.1} &  \textbf{57.4} &  20.1 &  83.2 & 30.3 &  16.0 &  44.5 &  43.6 & 51.5 \\ 
       Ours (Ensemble) & \textbf{88.7} & \textbf{41.6}	& 80.3 & 1.0 & 0.7 & 23.6 & 14.3 &  \textbf{33.1} & 81.9 & 81.1 & 57.2 & 21.1 & \textbf{84.1} & \textbf{33.4} & \textbf{19.1} & 44.3  & \textbf{44.1} & \textbf{52.3} \\ \hline
       \multicolumn{18}{c}{ResNet101 backbone} \\\hline
       AdvEnt\cite{vu_2019_advent}             & 85.6 & 42.2 & 79.7 & - & - & - & 5.4 & 8.1 & 80.4 & 84.1 & 57.9 & 23.8 & 73.3 & 36.4 & 14.2 & 33.0 & - & 48.0 \\
       LTIR \cite{Kim_2020_textureInv}         & 92.6 & 53.2 & 79.2 & - & - & -  & 1.6 & 7.5 & 78.6 & 84.4 & 52.6 & 20.0 & 82.1 & 34.8 & 14.6 & 39.4 & - & 49.3 \\
       BDL \cite{Li_19_BDL}                    & 86.0 & 46.7 & 80.3 & - & - & - & 14.1 & 11.6 & 79.2 & 81.3 & 54.1 & 27.9 & 73.7 & 42.2 & 25.7 & 45.3 & - & 51.4 \\
       FDA-MBT \cite{yang_2020_FDA}            & 79.3 & 35.0 & 73.2 & - & - & - & 19.9 & 24.0 & 61.7 & 82.6 & 61.4 & \textbf{31.1} & 83.9 & 40.8 & \textbf{38.4} & 51.1 & - & 52.5 \\ 
       PCEDA \cite{yang_20_PCEDA}              & 85.9 & 44.6 & 80.8 & - & - & - & 24.8 & 23.1 & 79.5 & 83.1 & 57.2 & 29.3 & 73.5 & 34.8 &  32.4 & 48.2 & - & 53.6 \\
       TPLD \cite{two_phase_ps}               & 80.9 & 44.3 & 82.2 & 19.9 & 0.3 & \textbf{40.6} & 20.5 & 30.1 & 77.2 & 80.9 & 60.6 & 25.5 & 84.8 & 41.1 & 24.7 & 43.7 & 47.3 & 53.5 \\
       Wang et al. \cite{wang_2021_uncertainty} & 79.4 & 34.6 & \textbf{83.5} & \textbf{19.3} & \textbf{2.8} & 35.3 & \textbf{32.1} & \textbf{26.9} & 78.8 & 79.6 & \textbf{66.6} & 30.3 & 86.1 & 36.6 & 19.5 & \textbf{56.9} & 48.0 & 54.6 \\
       PixMatch \cite{melas_2021_pixmatch} & \textbf{92.5} & \textbf{54.6} & 79.8 & 4.7 & 0.08 & 24.1 & 22.8 & 17.8 & 79.4 & 76.5 & 60.8 & 24.7 & 85.7 & 33.5 & 26.4 & 54.4 & 46.1 & 54.5 \\
       DPL-Dual (Ensemble) \cite{Cheng_21_dual_path}  & 87.5 & 45.7 & 82.8 & 13.3 & 0.6 & 33.2 & 22.0 & 20.1 & 83.1 & 86.0 & 56.6 & 21.9 & 83.1 & 40.3 & 29.8 & 45.7 & 47.0 & 54.2 \\ 
       SUDA \cite{SUDA} & 83.4 & 36.0 & 71.3 & 8.7 & 0.1 & 26.0 & 18.2 & 26.7 & 72.4 & 80.2 & 58.4 & 30.8 & 80.6 & 38.7 & 36.1 & 46.1 & 44.6 & 52.2 \\
       CaCo \cite{caco} & 87.4 & 48.9 & 79.6 & 8.8 & 0.2 & 30.1 & 17.4 & 28.3 & 79.9 & 81.2 & 56.3 &  24.2 & 78.6 & 39.2 & 28.1 & 48.3 & 46.0 & 53.6 \\ \hline
       Ours	& 85.8 & 41.7	& 82.4	& 7.6	& 1.9	& 33.2	& 26.5	& 18.4	& 83.3	& 86.5	& 62.0	& 29.7	& 83.9	& 52.1	& 34.6	& 51.4	& 48.8	& 56.8 \\ 
       Ours	(Ensemble) & 87.2 & 44.1	& 82.1 & 6.5 & 1.4 & 33.1 & 24.7 & 17.9	& \textbf{83.4} & \textbf{86.6} & 62.4 & 30.4	&  \textbf{86.1}	& \textbf{58.5} & 36.8 & 52.8 & \textbf{49.6}	& \textbf{57.9} \\ \hline

\end{tabular}}
\end{center}
\caption{{Quantitative comparison on SYNTHIA$\rightarrow$Cityscapes.}  We present per-class IoU and mean IoU (mIoU) obtained using VGG and ResNet101 backbones. mIoU and mIoU* are the mean IoU computed on the 16 classes and the 13 subclasses respectively.}
\label{tab:benchmark_syn2city_sup}
\end{table*}

\noindent \textbf{Training objective of the source domain network.}\\
As we mentioned in the main paper, for the source domain network we observed experimentally that we obtained better results by adding an entropy-based adversarial loss $\mathcal{L}_{adv}$, to the output of source-domain network $F_s$ when it is driven by translated target images. In Table \ref{tab:ent_loss} we report results obtained with and without the entropy-based adversarial loss (Eq. 9 in the main paper). Adding the  entropy-based regularization improves performance for most classes.

\begin{table}[!ht]
\begin{center}
\resizebox{0.94\textwidth}{!}{
\begin{tabular}{ccccccccccccccccccccc} \\ \hline
                 {Loss} & {\rotatebox[origin=l]{70}{road}} 
                          & {\rotatebox[origin=l]{70}{sidewalk}} 
                          & {\rotatebox[origin=l]{70}{building}} 
                          & {\rotatebox[origin=l]{70}{wall}} 
                          & {\rotatebox[origin=l]{70}{fence}} 
                          & {\rotatebox[origin=l]{70}{pole}} 
                          & {\rotatebox[origin=l]{70}{light}} 
                          & {\rotatebox[origin=l]{70}{sign}} 
                          & {\rotatebox[origin=l]{70}{vegetation}} 
                          & {\rotatebox[origin=l]{70}{terrain}} 
                          & {\rotatebox[origin=l]{70}{sky}} 
                          & {\rotatebox[origin=l]{70}{person}} 
                          & {\rotatebox[origin=l]{70}{rider}} 
                          & {\rotatebox[origin=l]{70}{car}} 
                          & {\rotatebox[origin=l]{70}{truck}} 
                          & {\rotatebox[origin=l]{70}{bus}} 
                          & {\rotatebox[origin=l]{70}{train}} 
                          & {\rotatebox[origin=l]{70}{motocycle}} 
                          & {\rotatebox[origin=l]{70}{bicycle}} 
                          & {mIoU} \\  \hline
$\mathcal{L}_{CE}$ &  90.2 &	38.0 &	81.2 &	 \textbf{29.1} &	16.2 &	 \textbf{24.4} &	23.7 &	15.5 &  \textbf{84.0}	&  \textbf{38.8} &	78.5 &	56.9 &	24.0	&  85.0 &	 \textbf{36.4}	&  \textbf{47.0} &	0.3 &	 \textbf{31.8}	& 26.8 & 43.6 \\ \hline
$\mathcal{L}_{CE} + \mathcal{L}_{adv}$          &  \textbf{90.5}  & \textbf{ 39.4} &	\textbf{82.0} &	29.0 &  \textbf{21.4} &	23.6 &	 \textbf{28.6} &	 \textbf{17.8} &	83.9 &	38.2 &	 \textbf{79.8} &	 \textbf{56.9}	&  \textbf{26.0} &	 \textbf{85.1} &	32.2 &	44.1 &	 \textbf{3.8}	& 31.5	&  \textbf{30.1} &  \textbf{44.4}     \\ \hline
\end{tabular}}
\end{center}
\caption{Better performance is achieved by adding an entropy-based regularization $\mathcal{L}_{adv}$ to the output of source-domain network $F_s$ when it is driven by translated target images.}
\label{tab:ent_loss}
\end{table}

\noindent \textbf{Selection of $r$ for pseudo-label generation}.\\
As we mentioned in \refsec{sec:pseudo}, we select the proportion of $r$ based on the mIoU of the validation set. In Table~\ref{tab:sel_r} and Table~\ref{tab:sel_r_round_2} we provide the per-class IoU and mean IoU (mIoU) obtained on the validation set for different values of $r$ in the first (R=0) and second (R=1) round of pseudo-labeling respectively. In both rounds the best performance is achieved for $r=0.6$. In the second round of pseudo-labeling the networks provide more confident predictions since the performance remains the same for almost all classes when $r\leq0.5$.

\begin{table}[!ht]
\begin{center}
\resizebox{0.94\textwidth}{!}{
\begin{tabular}{ccccccccccccccccccccc} \\ \hline
                 {$r$} & {\rotatebox[origin=l]{70}{road}} 
                          & {\rotatebox[origin=l]{70}{sidewalk}} 
                          & {\rotatebox[origin=l]{70}{building}} 
                          & {\rotatebox[origin=l]{70}{wall}} 
                          & {\rotatebox[origin=l]{70}{fence}} 
                          & {\rotatebox[origin=l]{70}{pole}} 
                          & {\rotatebox[origin=l]{70}{light}} 
                          & {\rotatebox[origin=l]{70}{sign}} 
                          & {\rotatebox[origin=l]{70}{vegetation}} 
                          & {\rotatebox[origin=l]{70}{terrain}} 
                          & {\rotatebox[origin=l]{70}{sky}} 
                          & {\rotatebox[origin=l]{70}{person}} 
                          & {\rotatebox[origin=l]{70}{rider}} 
                          & {\rotatebox[origin=l]{70}{car}} 
                          & {\rotatebox[origin=l]{70}{truck}} 
                          & {\rotatebox[origin=l]{70}{bus}} 
                          & {\rotatebox[origin=l]{70}{train}} 
                          & {\rotatebox[origin=l]{70}{motocycle}} 
                          & {\rotatebox[origin=l]{70}{bicycle}} 
                          & {mIoU} \\  \hline

1.0 & 92.1  &	47.8  &	84.3  &	36.5  &	27.9 &	31.5 &	36.6 &	24.5  &	85.4 &	41.2 &  81.6 &	61.4 &	30.1 &	86.3    &   37.6 &	47.3 &	1.3 &	28.7 &	32.7 &	48.2 \\ \hline
0.8	& 97.4	&   69.8  & 93.4. & 52.0  &	42.5 &  47.1 &  55.2 &  37.7  & 94.2 &  53.5 &  91.1 &  78.2 &  40.2 &  94.4	&   47.4 &  55.7 &  2.2 &	40.9 &  55.4 &  60.4 \\ \hline
0.7 & 98.2  &  \textbf{72.1}  & 95.9. & 63.4. & 51.5 &  \textbf{55.2} & \textbf{64.0} & \textbf{42.5}   &	96.5 &	65.0 &  93.7 &	86.9 &  51.7 &	96.4    &	57.0 &	61.2 &	\textbf{2.7} &	53.1 & 	65.8 &	67.0 \\ \hline
0.6	& 98.5	&   70.2  &	96.3 & 	73.4  & \textbf{57.5} &	54.1 &	58.7 &	33.3  &	97.2 &	77.4 &	\textbf{93.8} &	90.8 &  60.3 &	97.5    &	65.5 &	69.6 &	2.5	&   65.8 &  \textbf{71.9} &	\textbf{70.2} \\ \hline
0.5	& \textbf{98.6}	&   59.4  &	96.5 &  \textbf{79.0}  & 55.2 &	44.6 &	54.3 &	20.7  &	97.6 &	81.4 &	93.8 &	92.2 &	61.0 &  \textbf{97.9}    &	\textbf{70.6} &	\textbf{74.1} &	1.3 &	\textbf{73.3} &	70.3 &  69.6 \\ \hline
0.4	& 98.6  &	50.1  &	\textbf{96.6} &	76.5  &	42.2 &	44.6 &	54.6 &	20.6  & \textbf{97.7} & \textbf{82.3} &  93.8 &	\textbf{92.3} &	\textbf{61.1} &	97.9	&   70.6 &	74.1 &	0.5 &	73.4 &	70.5 &	68.3 \\ \hline
0.3	& 98.6  &	50.1  &	96.6 &	76.6  &	38.3 &	44.7 &	54.6 &	20.6  &	97.7 &	82.3 &	93.8 &	92.3 &	61.1 &	97.9	&   70.6 &	74.1 &	0.2 & 	73.4 &	70.5 &	68.1 \\ \hline
\end{tabular}}
\end{center}
\caption{Per-class IoU and mean (mIoU) obtained using different values of $r$ for class-wise confidence threshold selection in the first round (R=0) of pseudo-labeling. We observe that $r=0.6$ gives the best results.}
\label{tab:sel_r}
\end{table}

\begin{table}[!ht]
\begin{center}
\resizebox{0.94\textwidth}{!}{
\begin{tabular}{ccccccccccccccccccccc} \\ \hline
                 {$r$} & {\rotatebox[origin=l]{70}{road}} 
                          & {\rotatebox[origin=l]{70}{sidewalk}} 
                          & {\rotatebox[origin=l]{70}{building}} 
                          & {\rotatebox[origin=l]{70}{wall}} 
                          & {\rotatebox[origin=l]{70}{fence}} 
                          & {\rotatebox[origin=l]{70}{pole}} 
                          & {\rotatebox[origin=l]{70}{light}} 
                          & {\rotatebox[origin=l]{70}{sign}} 
                          & {\rotatebox[origin=l]{70}{vegetation}} 
                          & {\rotatebox[origin=l]{70}{terrain}} 
                          & {\rotatebox[origin=l]{70}{sky}} 
                          & {\rotatebox[origin=l]{70}{person}} 
                          & {\rotatebox[origin=l]{70}{rider}} 
                          & {\rotatebox[origin=l]{70}{car}} 
                          & {\rotatebox[origin=l]{70}{truck}} 
                          & {\rotatebox[origin=l]{70}{bus}} 
                          & {\rotatebox[origin=l]{70}{train}} 
                          & {\rotatebox[origin=l]{70}{motocycle}} 
                          & {\rotatebox[origin=l]{70}{bicycle}} 
                          & {mIoU} \\  \hline

1.0 &   93.0 &	53.3 & 	85.8 &	41.2 &	33.1 & 	33.4 &	39.1 &	29.7 &	86.4 &	45.4 & 84.5 &	60.0  &	29.3 &	86.9 &	45.8 &	57.7	& 2.7 &	34.6 & 45.8	& 52.0  \\ \hline
0.8 &   96.8	&  \textbf{67.4} &	93.6 &	57.3 &	45.2	& 46.7 &	 \textbf{54.2} &	 \textbf{41.9} &	94.2 &	59.2	& 92.4 &	74.8 &	37.1 &	94.4	& 57.1 & 71.3	& 4.6	& 49.2	& 60.8 & 63.1 \\ \hline
0.7 &   97.2	& 67.3 & 	94.2 &	64.5 &	49.4 &	 \textbf{49.0}	& 51.9	& 37.1 & 	95.3 &	67.6 &	 \textbf{92.5} &	80.8 &	46.3 &	95.9	 & 68.6 &	77.9 &	 \textbf{5.4} & 	59.7 &	70.5 &	66.9 \\ \hline
0.6 &   \textbf{97.3} &	65.4 &	 \textbf{94.5} &	67.8 &	 \textbf{51.3} &	44.0	& 48.5	& 36.0 &	 \textbf{95.7} &	71.9 &	92.5 &	84.4 &	52.5 &	96.9 &	85.3 &	81.2 &	4.9	& 67.9 &	74.7 &	69.1 \\ \hline
0.5 &	97.3 &	65.4 &	94.5 &	 \textbf{68.1} &	49.3 &	42.8 &	48.6 &	36.3 &	95.7 &	 \textbf{72.2}	& 92.5 &	 \textbf{84.7} &	 \textbf{52.7} &	 \textbf{97.0} &	 \textbf{87.5} 	&  \textbf{81.6} & 	3.0 &	 \textbf{68.6} &	 \textbf{75.1} &	 \textbf{69.1}	\\ \hline
0.4 &	97.3 &	65.4 &	94.5 &	68.1 &	49.3 &	42.8 &	48.6	& 36.3 &	95.7 &	72.2	& 92.5 &	84.7 &	52.7 &	97.0 &	87.5	& 81.6 &	1.3 &	68.6 &	75.1 &	69.0 \\ \hline
0.3 &	97.3 &	65.4 &	94.5 &	68.1 &	49.3 &	42.8 &	48.6 &	36.3 &	95.7 &	72.2	& 92.5 & 	84.7 &	52.7 &	97.0 &	87.5 &	81.6 &	0.4 &	68.6	& 75.1 &	69.0 \\ \hline
\end{tabular}}
\end{center}
\caption{Per-class IoU and mean (mIoU) obtained using different values of $r$ for class-wise confidence threshold selection in the second round (R=1) of pseudo-labeling. We observe that $r=0.6$ gives the best results.}
\label{tab:sel_r_round_2}
\end{table}

\noindent \textbf{Class-wise IoU for ablation studies reported in the main paper}.\\
In Table~\ref{tab:stoch_vs_det} we report the per-class IoU obtained from deterministic and stochastic translation (mIoU results are reported in Table 1 of the main paper). In particular the comparison builds on directly on the ADVENT baseline\cite{vu_2019_advent}; the first two rows compare the originally published and our reproduced numbers respectively. The third row shows the substantial improvement attained by training the system of ADVENT using translated images. The forth row reports our stochastic translation-based result. We observe a substantial improvement, that can be attributed solely to the stochasticity of the translation. The last row shows that imposing a semantic consistency constraint as described in Eq. 5 (main paper) further improves the performance. In Table~\ref{tab:ens_gta_cityscapes_2_iters_sup} we report the per-class IoU obtained from multiple rounds $R$ of pseudo-labeling and training (mIoU results are provided in Table 3 of the main paper). Multiple rounds of pseudo-labeling and training yield improved performance.
\begin{table*}[ht]
\begin{center}
\resizebox{0.94\textwidth}{!}{
\begin{tabular}{ccccccccccccccccccccc} \\ \hline
                 {Model} & {\rotatebox[origin=l]{70}{road}} 
                          & {\rotatebox[origin=l]{70}{sidewalk}} 
                          & {\rotatebox[origin=l]{70}{building}} 
                          & {\rotatebox[origin=l]{70}{wall}} 
                          & {\rotatebox[origin=l]{70}{fence}} 
                          & {\rotatebox[origin=l]{70}{pole}} 
                          & {\rotatebox[origin=l]{70}{light}} 
                          & {\rotatebox[origin=l]{70}{sign}} 
                          & {\rotatebox[origin=l]{70}{vegetation}} 
                          & {\rotatebox[origin=l]{70}{terrain}} 
                          & {\rotatebox[origin=l]{70}{sky}} 
                          & {\rotatebox[origin=l]{70}{person}} 
                          & {\rotatebox[origin=l]{70}{rider}} 
                          & {\rotatebox[origin=l]{70}{car}} 
                          & {\rotatebox[origin=l]{70}{truck}} 
                          & {\rotatebox[origin=l]{70}{bus}} 
                          & {\rotatebox[origin=l]{70}{train}} 
                          & {\rotatebox[origin=l]{70}{motocycle}} 
                          & {\rotatebox[origin=l]{70}{bicycle}} 
                          & {mIoU} \\  \hline

                ADVENT & 89.9 & 36.5 & 81.6 & 29.2 & 25.2 & 28.5 & 32.3 & 22.4 & 83.9 & 34.0 & 77.1 & 57.4 & 27.9 & 83.7 & 29.4 & 39.1 & 1.5 & 28.4 & 23.3 & 43.8  \\ \hline

                ADVENT $^{\ast}$ & 87.2	& 38.5	& 78.2 & 25.9	& 24.6	& 30.4	& 36.3	& 21.7	& 84.0	& 28.7	& 76.7	& 60.1	& 28.8	& 80.0	& 28.0	& 45.2	& 0.7	& 19.7	& 19.9	& 42.9 \\ \hline

                ADVENT $^{\ast}+$ \\ CycleGAN$^{\ast}$ & \textbf{91.9} &	\textbf{51.5} &	83.1 &	30.8 &	23.6 &	\textbf{32.0} &	32.1 &	24.3 &	83.8 &	\textbf{38.5} &	\textbf{82.3} &	58.7 &	28.5 &	84.1 &	33.3 & 	35.9 &	0.6	 & 21.7 &	20.0 &	45.1  \\ \hline
                
                Ours & 90.2	& 37.6 & 	\textbf{84.1} & \textbf{33.0} &	\textbf{25.1} &	30.1 &	36.8 &	\textbf{28.4} &	83.8 &	36.1 &	82.2 &	58.1 &	\textbf{29.6}	& 84.6 &	\textbf{34.4} &	45.4 &	1.0 &	\textbf{26.2} &	\textbf{30.8} &	46.2 \\ \hline
                
                Ours w/ $L_{sem}$  & 92.1 & 49.9 &	83.5  & 29.1	& 24.7	& 30.3	& \textbf{38.3}	& 27.2	& \textbf{84.8}	& 34.4	& 81.1	& \textbf{60.4}	& 28.1	& \textbf{85.2}	& 33.0	& \textbf{45.7}	& \textbf{2.5}	& 23.8	& 30.4	& \textbf{46.6} \\ \hline

\end{tabular}}
\end{center}
\caption{GTA to Cityscapes UDA using stochastic translation: We train ADVENT using synthetic images obtained from deterministic translation (CycleGAN) and stochastic translation (Ours). We observe a clear improvement thanks to pixel-space alignment based on stochastic translation. $^{\ast}$ denotes our retrained models.}
\label{tab:stoch_vs_det}
\end{table*}

\begin{table*}[ht]
\begin{center}
\resizebox{0.94\textwidth}{!}{
\begin{tabular}{ccccccccccccccccccccc} \\ \hline
                 {Model} & {\rotatebox[origin=l]{70}{road}} 
                          & {\rotatebox[origin=l]{70}{sidewalk}} 
                          & {\rotatebox[origin=l]{70}{building}} 
                          & {\rotatebox[origin=l]{70}{wall}} 
                          & {\rotatebox[origin=l]{70}{fence}} 
                          & {\rotatebox[origin=l]{70}{pole}} 
                          & {\rotatebox[origin=l]{70}{light}} 
                          & {\rotatebox[origin=l]{70}{sign}} 
                          & {\rotatebox[origin=l]{70}{vegetation}} 
                          & {\rotatebox[origin=l]{70}{terrain}} 
                          & {\rotatebox[origin=l]{70}{sky}} 
                          & {\rotatebox[origin=l]{70}{person}} 
                          & {\rotatebox[origin=l]{70}{rider}} 
                          & {\rotatebox[origin=l]{70}{car}} 
                          & {\rotatebox[origin=l]{70}{truck}} 
                          & {\rotatebox[origin=l]{70}{bus}} 
                          & {\rotatebox[origin=l]{70}{train}} 
                          & {\rotatebox[origin=l]{70}{motocycle}} 
                          & {\rotatebox[origin=l]{70}{bicycle}} 
                          & {mIoU} \\  \hline
                
                source (R=0) &  90.5  & 39.4 &	82.0 &	29.0 & 21.4 &	23.6 &	28.6 &	17.8 &	83.9 &	38.2 &	79.8 &	56.9	& 26.0 &	85.1 &	32.2 &	44.1 &	3.8	& 31.5	& 30.1 & 44.4  \\

                target, $\sigma^2=1$ (R=0) & 92.1 & 49.9 &	83.5  & 29.1	& 24.7	& 30.3	& 38.3	& 27.2	& 84.8	& 34.4	& 81.1	& 60.4	& 28.1	& 85.2	& 33.0	& 45.7	& 2.5	& 23.8	& 30.4	& 46.6 \\

                target, $\sigma^2=10$ (R=0) & 90.9 &	43.0 &	83.4	& 30.6	& 29.3 &	30.6	& 34.1 &	27.1 &	84.4	& 36.2	& 79.9	& 60.6	& 29.5	& 84.5	& 32.5	& 40.3	& 3.1	& 29.2	& 26.4	& 46.1 \\
                
                Ens (R=0) &   92.1  &	47.8  &	84.3  &	36.5  &	27.9  &	31.5  &	36.6  &	24.5  &	85.4  &	41.2  &	81.6  &	61.4  &	30.1  &	86.3  &	37.6 &	47.3 &	1.3 &	28.7 &	32.7 &	48.2   \\  \hline

                source (R=1) &  92.1	& 48.4 &	84.3	& 36.4 & 	29.5 &	30.5 &	35.9 &	26.5 & 85.4 &	42.9 & 	82.1  & 59.8 &	29.6 &	85.5	& 38.2 &	52.9 &	3.4 &	32.7 &	37.3 &	49.1  \\

                target, $\sigma^2=1$ (R=1) & 92.1 & 47.5 &	85.1 &	38.3 &	29.4 &	32.9 &	35.4 &	32.1 &	85.9 &	46.8 &	81.7 &	60.5 &	30.4	& 86.6 &	35.7 &	51.1 &	4.4	& 34.9 &	41.0	& 50.1 \\

                target, $\sigma^2=10$ (R=1) & 92.9 &	55.2 &	85.1 &	38.1 &	30.6 &	32.8 &	39.8 &	34.8 &	85.9 &	42.2 &	84.0	& 59.0 &	26.1 &	85.4 &	47.9 &	46.3 &	10.1 & 28.4	& 42.8	& 50.9 \\
                
                Ens (R=1) &   93.0 &	53.3 & 	85.8 &	41.2 &	33.1 & 	33.4 &	39.1 &	29.7 &	86.4 &	45.4 & 84.5 &	60.0  &	29.3 &	86.9 &	45.8 &	57.7	& 2.7 &	34.6 & 45.8	& 52.0   \\  \hline

                source (R=2) &  92.3 & 48.2	& 85.1	& 40.7	& 34.3	& 29.8	& 38.5	& 28.2	& 86.5	& 46.7	& 83.3	& 60.9	& 30.2	& 86.9	& 41.3	& 53.1	& 10.4	& 38.4	& 40.5	& 51.3  \\

                target, $\sigma^2=1$ (R=2) & 93.3	& 56.5	& 85.9	& 41.0	& 33.1	& 34.8	& 43.8	& 43.8	& 86.6	& 46.5	& 82.5	& 61.1	& 30.4 &	87.0	& 39.7	& 50.7	& 8.8 &	34.9 &	46.8 &	53.0 \\

                target, $\sigma^2=10$ (R=2) & 93.4	& 56.3 & 	85.6 &	40.6 & 	33.5 &	35.9 &	43.5 &	41.1 &	85.7 &	43.8 &	84.1 &	60.6 &	29.2 &	87.2 &	44.2 &	53.7 &	13.7 &	33.8 &	39.2 &	52.8 \\
                
                Ens (R=2) &    93.4 &    55.8 &    86.4 & 44.4 & 36.1 & 34.6 & 45.0 & 39.8 &    86.9 &   48.0 & 84.4 &  61.7 & 30.9 & 87.7 & 44.9 & 55.9 & 11.1 & 38.4 &   45.4 & 54.3   \\  \hline
\end{tabular}}
\end{center}
\caption{Ablation study on GTA$\rightarrow$Cityscapes. Averaging the predictions (Ens) of a source network $F_s$, and two target networks $F_t$ trained with different degrees of stochasticity ($\sigma^2$) in the translation allows to obtain robust pseudo-labels, while using multiple rounds R of
pseudo-labeling and training improves the overall performance.}
\label{tab:ens_gta_cityscapes_2_iters_sup}
\end{table*}

\subsubsection{Qualitative results}
\begin{figure*}[ht]
\centering
\includegraphics[width=\textwidth]{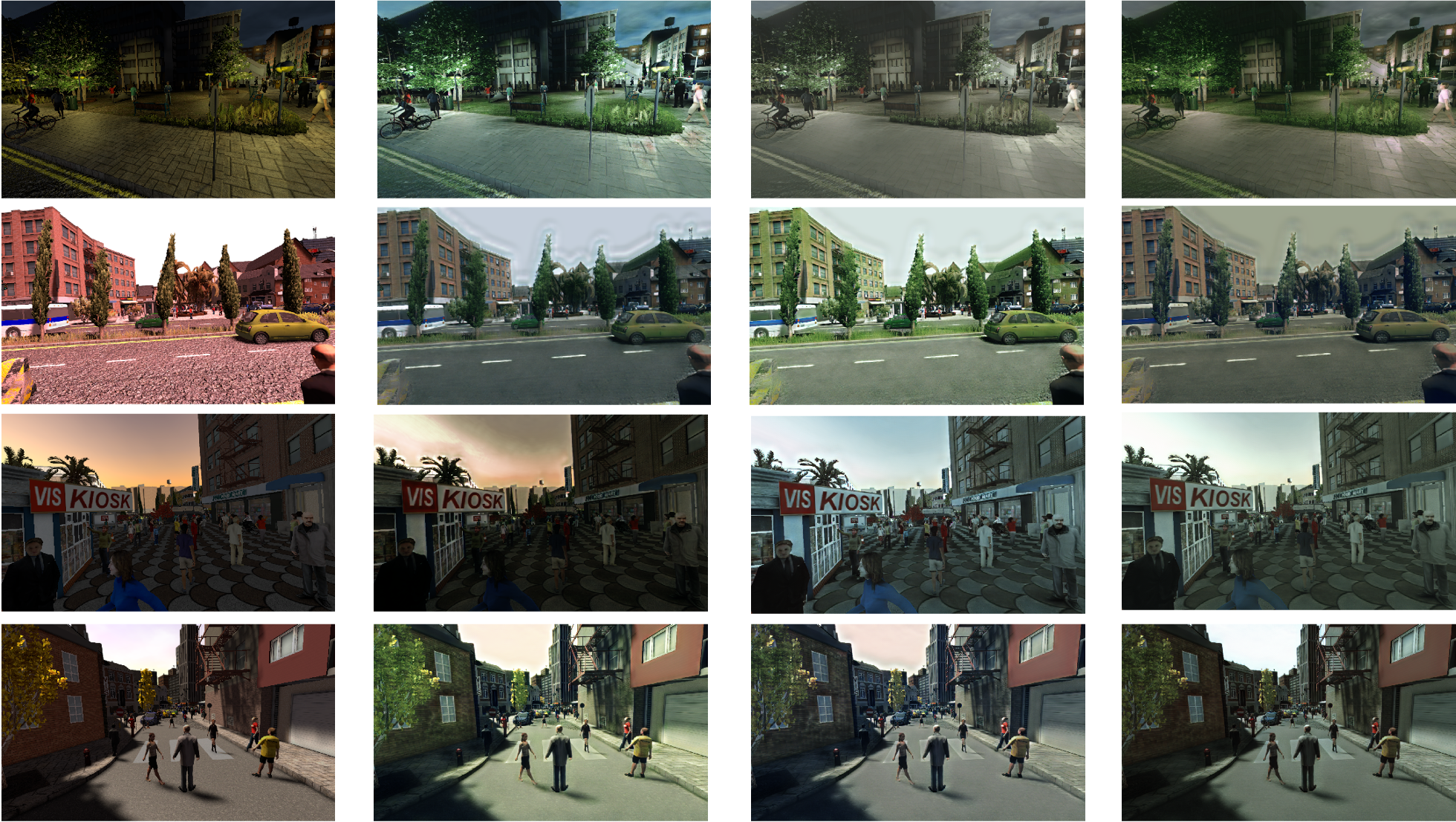}
\caption{Diverse translations of  images from the 
SYNTHIA 
source dataset to the Cityscapes target dataset: we observe that even though the content and pixel semantics stay intact, we generate diverse variants of the same scene, effectively capturing more faithfully the data distribution in the target domain. 
}
\label{fig:synthia_samples}
\end{figure*}

\begin{figure*}[h]
\centering
\includegraphics[width=\textwidth]{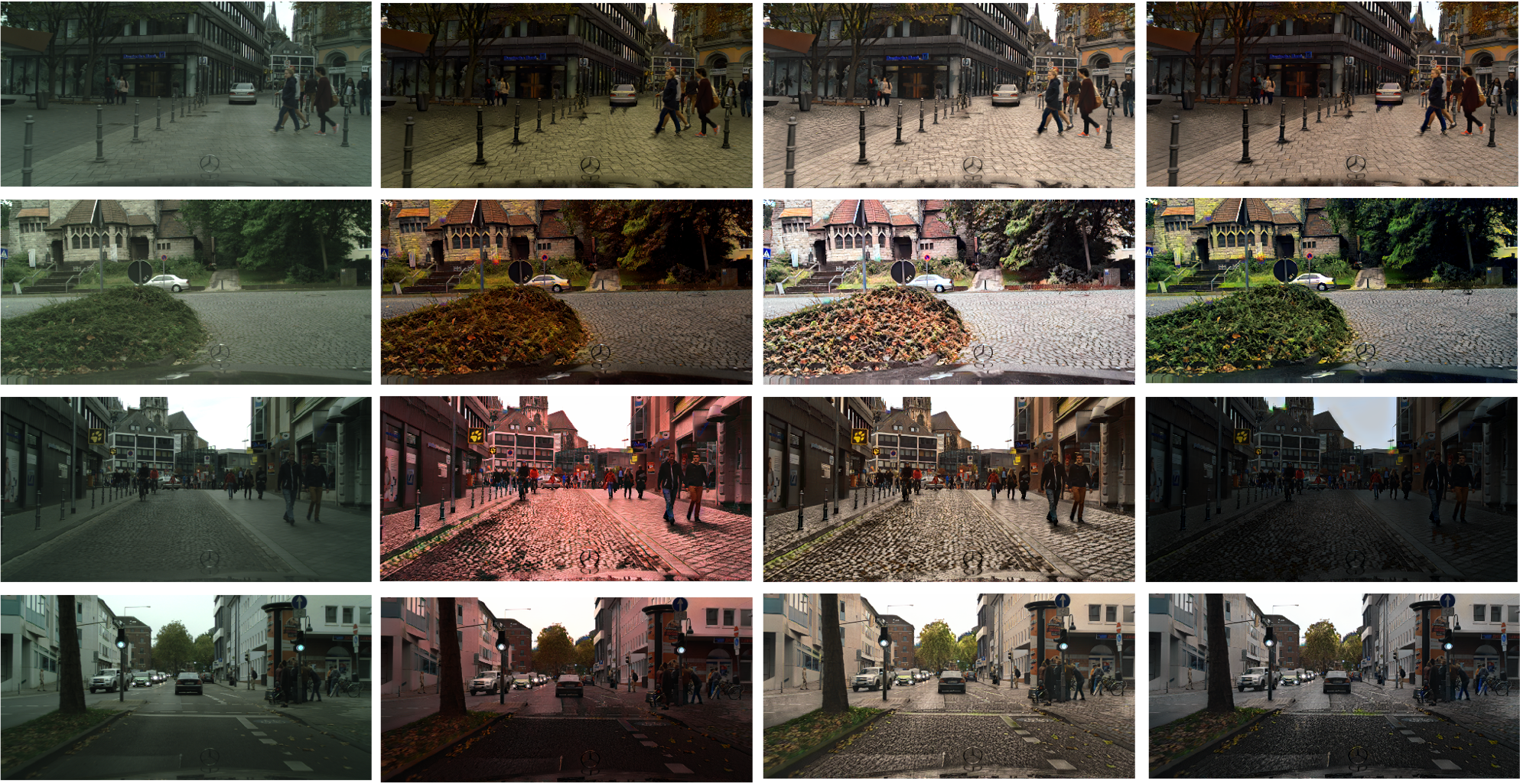}
\caption{Diverse translations of  images from the  
 Cityscapes target dataset to the SYNTHIA 
source dataset: we observe that even though the content and pixel semantics stay intact, we generate diverse variants of the same scene, effectively capturing more faithfully the data distribution of the source domain. This allows us to generate more robust pseudo-labels.  
}
\label{fig:cityscapes2synthia_samples}
\end{figure*}

\begin{figure*}[h]
\centering
\includegraphics[width=\textwidth]{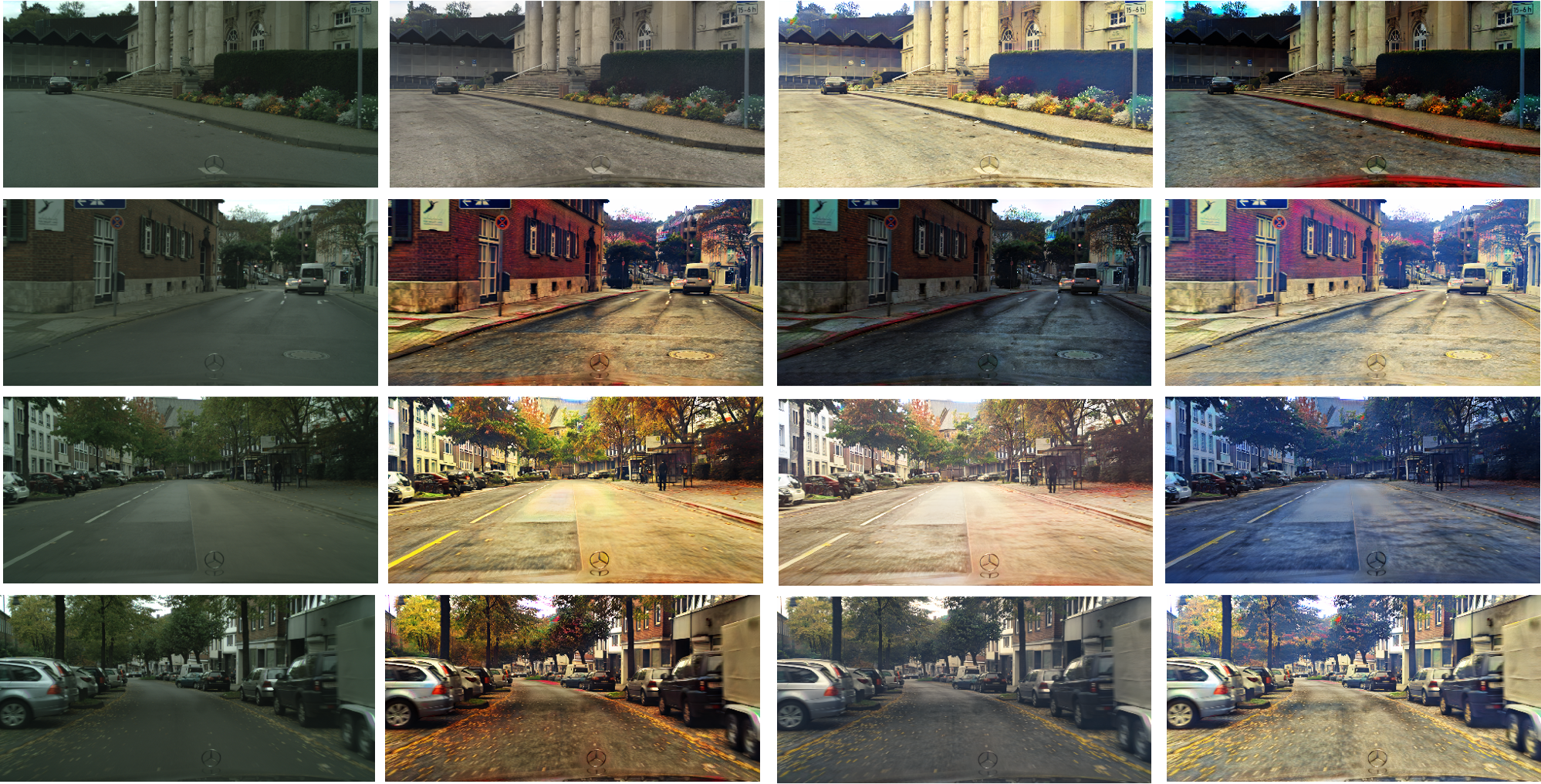}
\caption{Diverse translations of  images from the  
 Cityscapes target dataset to the GTA 
source dataset: we observe that even though the content and pixel semantics stay intact, we generate diverse variants of the same scene, effectively capturing more faithfully the data distribution of the source domain. This allows us to generate more robust pseudo-labels. 
}
\label{fig:cityscapes2gta_samples}
\end{figure*}

\textbf{Diverse translation obtained using stochastic translation}.\\
\noindent\reffig{fig:synthia_samples} shows diverse translations of  images from the SYNTHIA source dataset to the Cityscapes target dataset. \reffig{fig:cityscapes2synthia_samples} and \reffig{fig:cityscapes2gta_samples} show diverse translations of  images from the Cityscapes target dataset to the SYNTHIA and GTA source datasets respectively. We observe that stochastic translation generates diverse samples that capture more faithfully the data distribution of the source domain and preserve the content of the original image allowing us to obtain more robust pseudo-labels for the target data.


\begin{figure*}[h]
\centering
    \includegraphics[width=\textwidth]{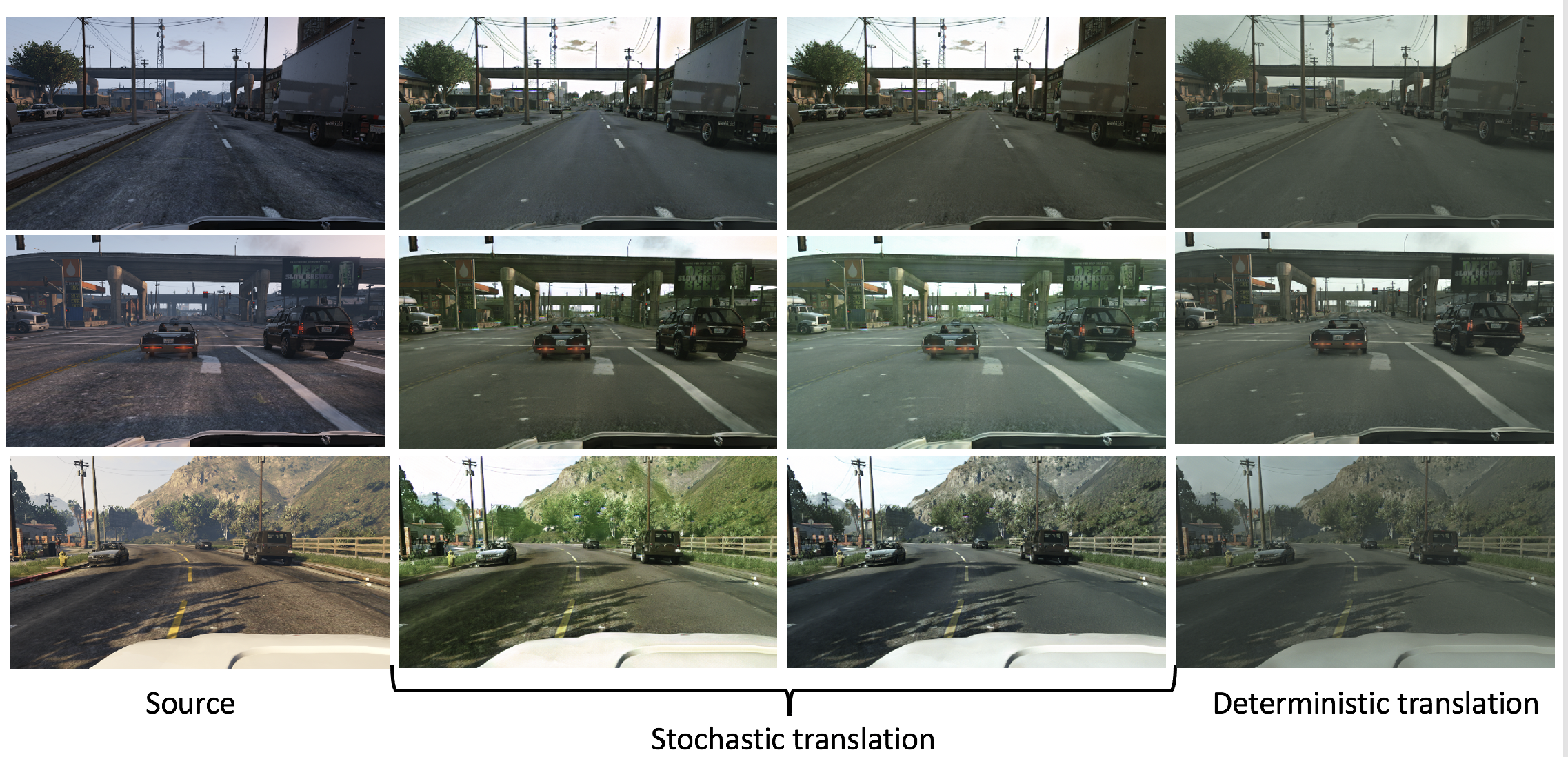}
\caption{Stochastic and deterministic translation of images from the GTA source dataset to the Cityscapes target dataset.}
\label{fig:gta2city_samples_svd}
\end{figure*}

\begin{figure*}[h]
\centering
    \includegraphics[width=\textwidth]{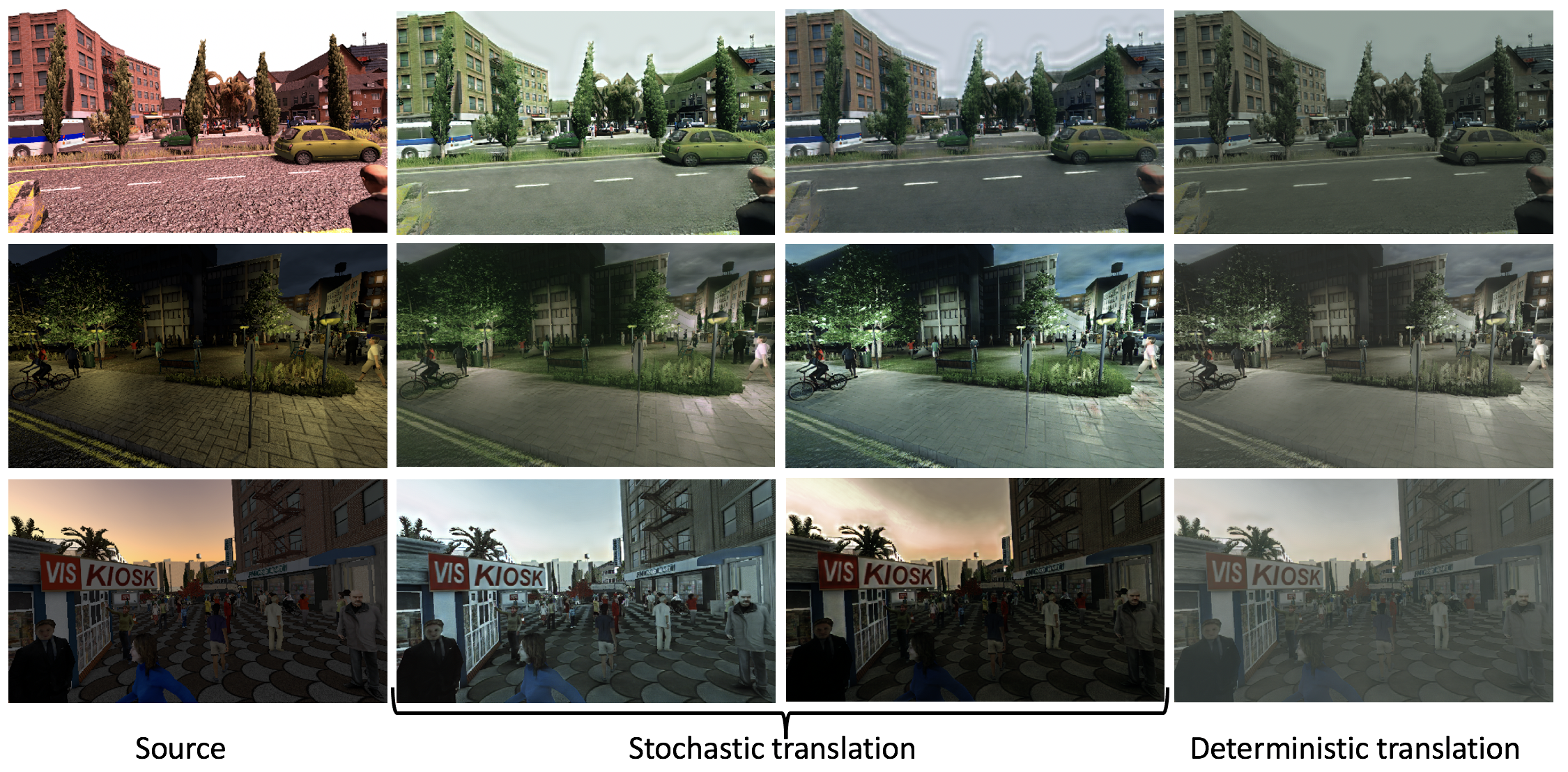}
\caption{Stochastic and deterministic translation of images from the SYNTHIA source dataset to the Cityscapes target dataset.}
\label{fig:sythia2city_samples_svd}
\end{figure*}

\begin{figure*}[h]
\centering
    \includegraphics[width=\textwidth]{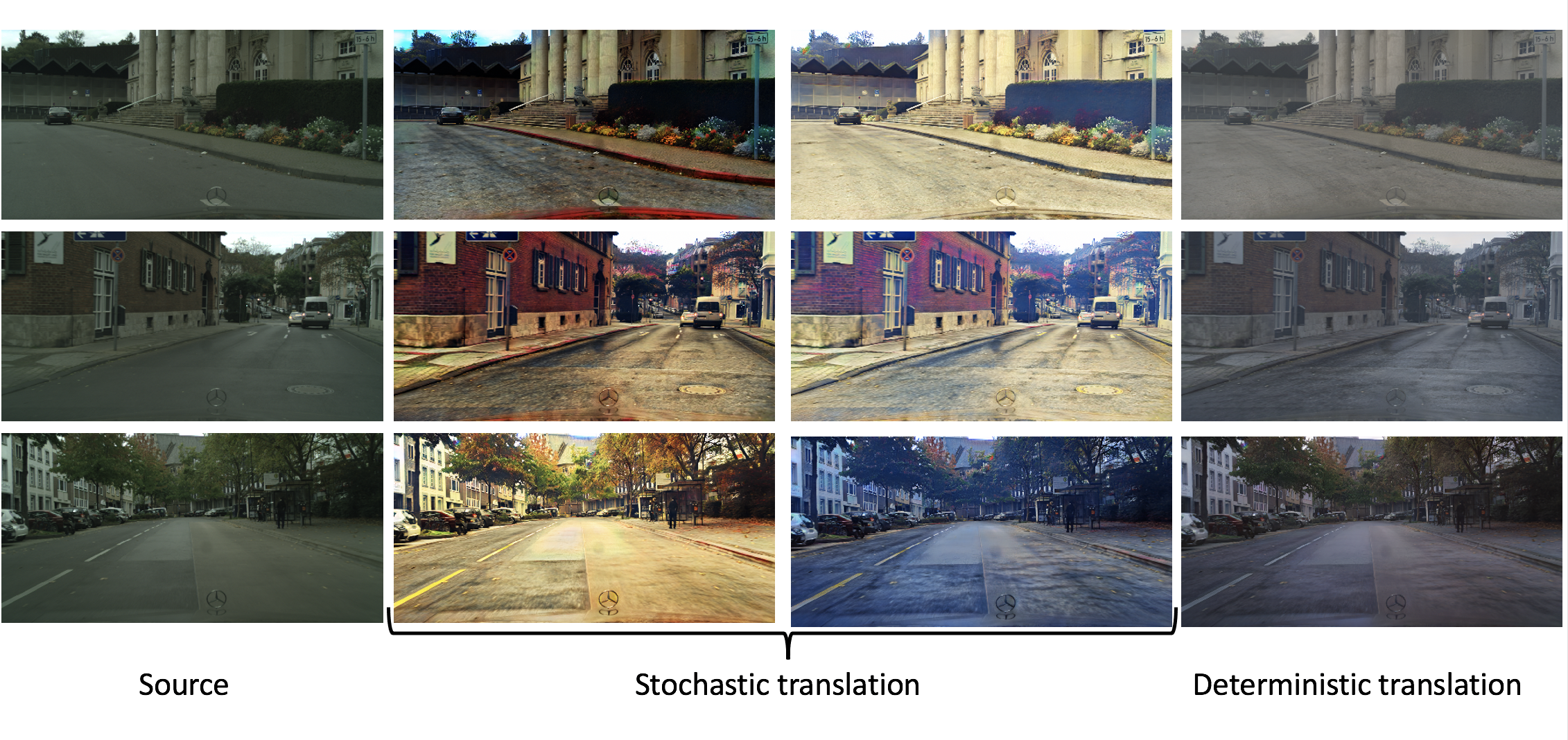}
\caption{Stochastic and deterministic translation of images from the Cityscapes target dataset to the GTA source dataset.}
\label{fig:city2gta_samples_svd}
\end{figure*}

\begin{figure*}[h]
\centering
    \includegraphics[width=\textwidth]{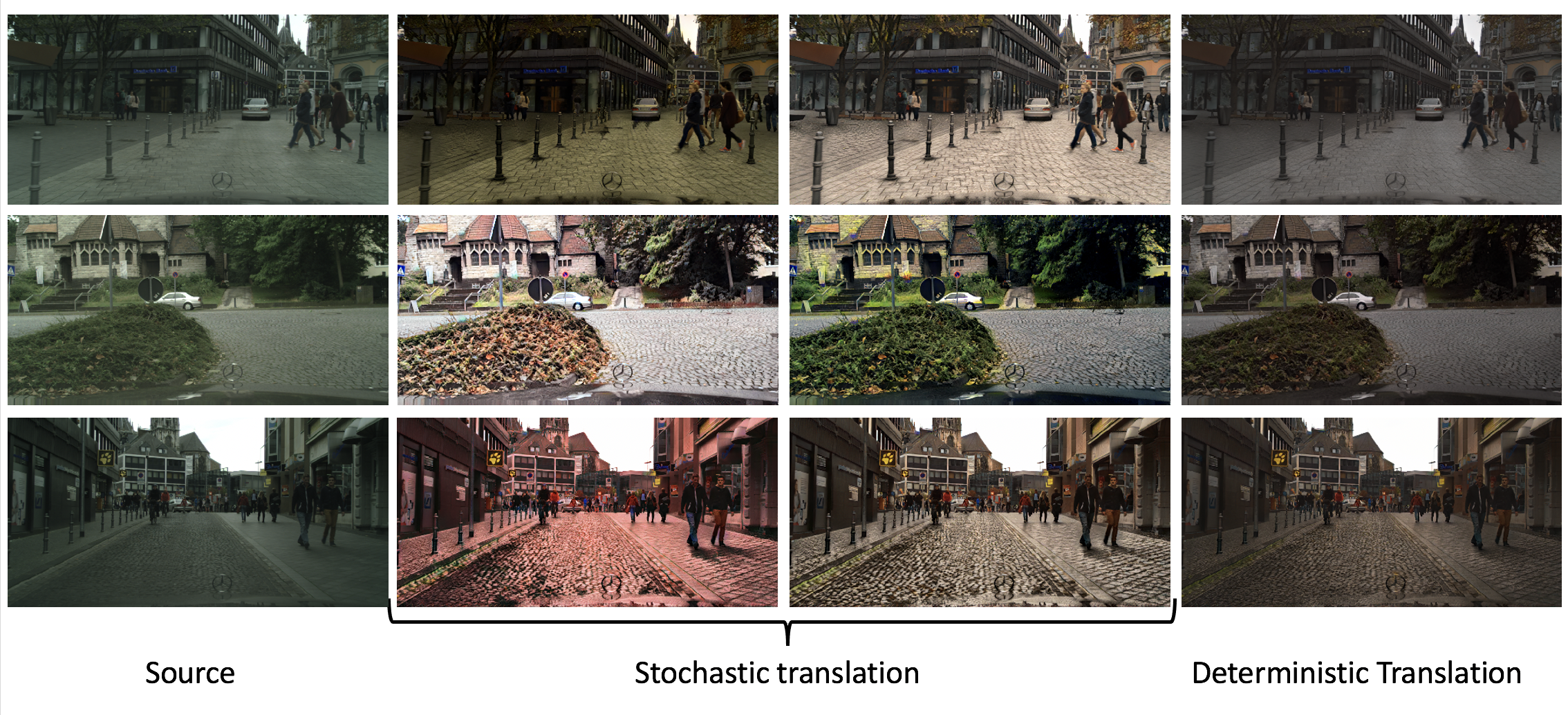}
\caption{Stochastic and deterministic translation of images from the Cityscapes target dataset to the SYNTHIA source dataset.}
\label{fig:city2sytnthia_samples_svd}
\end{figure*}

\noindent\textbf{Stochastic versus deterministic translation}. \\
\noindent\reffig{fig:gta2city_samples_svd} shows stochastic and deterministic translation of images from the GTA source dataset to the Cityscapes target dataset while \reffig{fig:sythia2city_samples_svd} shows stochastic and deterministic translation of images from the SYNTHIA source dataset to the Cityscapes target dataset. \noindent\reffig{fig:city2gta_samples_svd} shows stochastic and deterministic translation of images from the Cityscapes target dataset to the GTA source dataset  while \reffig{fig:city2sytnthia_samples_svd} shows stochastic and deterministic translation of images from the Cityscapes target dataset to the SYNTHIA source dataset.  We observe that stochastic translation generates sharp samples of noticeable diversity compared to the deterministic translation that generates a single output.

\begin{figure*}[h]
\centering
    \includegraphics[width=\textwidth]{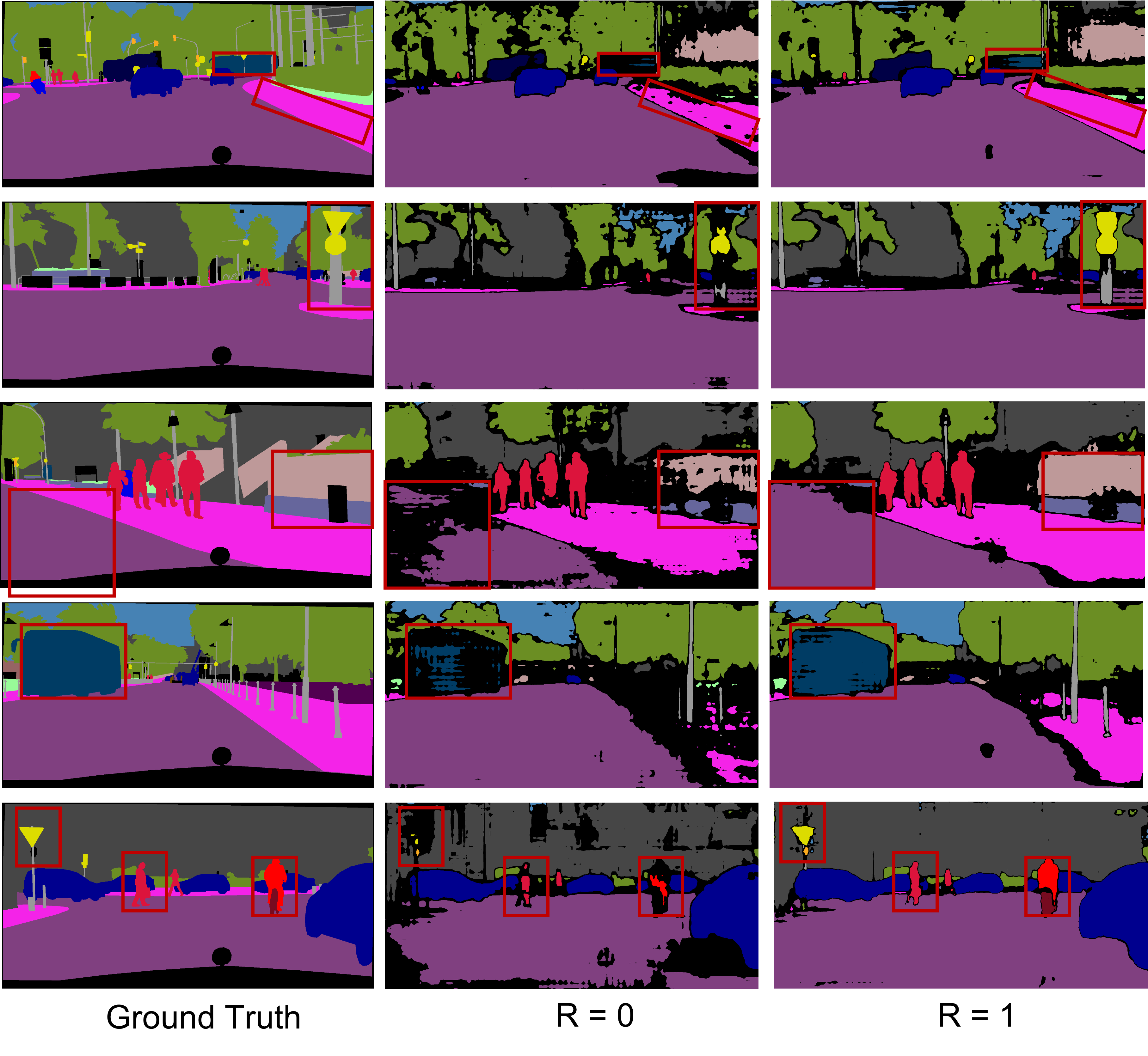}
\caption{Visualization of pseudo-labels obtained from the first (R=0) and second (R=1) round. Pseudo-labels  obtained in the second round are more accurate allowing us to train more robust models in the last round of training.}
\label{fig:pseudo_multi}
\end{figure*}

\noindent\textbf{Multiple rounds of pseudo-labeling}. \\
\noindent\reffig{fig:pseudo_multi} shows the pseudo-labels obtained from the first (R=0) and second (R=1) round of pseudo-labeling. We observe that the pseudo-labels we obtained in the second round are more accurate allowing us to train more accurate models in the last round of training. 

\begin{figure*}[h]
\centering
    \includegraphics[width=\textwidth]{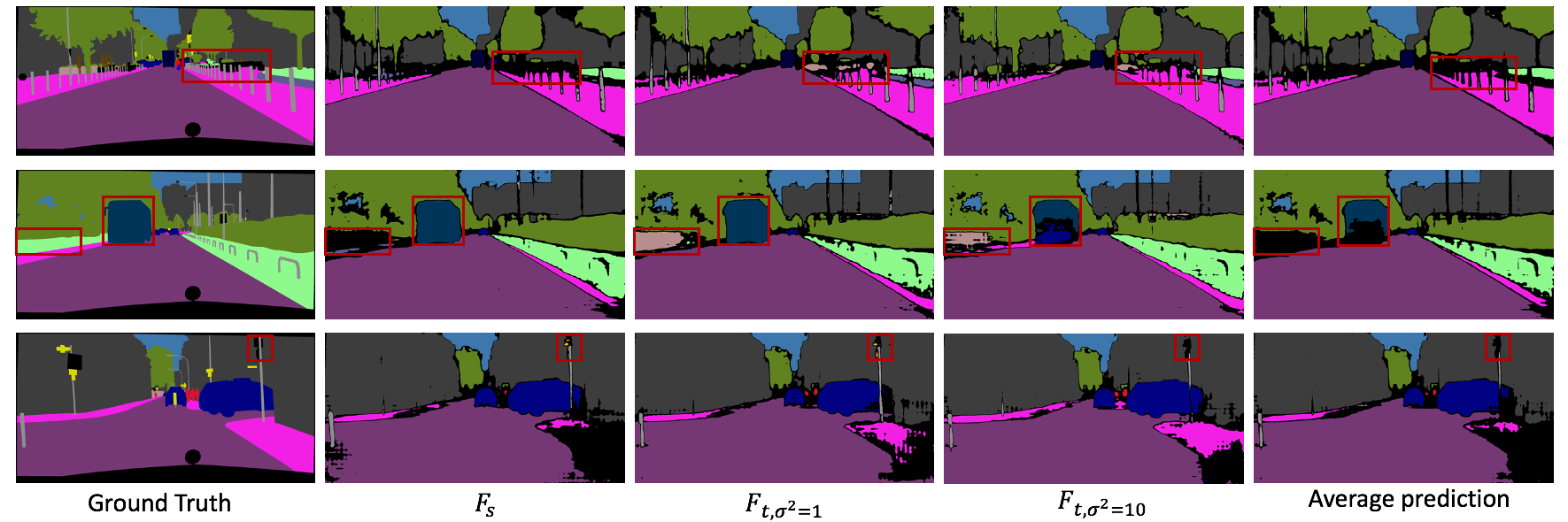}
\caption{Visualization of pseudo-labels obtained by averaging the predictions of two target networks $F_{t,\sigma^2=1}$, $F_{t,\sigma^2=10}$ and a one source network $F_s$. Averaging the predictions allows us to generate more accurate pseudo-labels.}
\label{fig:pseudo_ens}
\end{figure*}

\noindent\textbf{Robust pseudo-labeling through ensembling}. \\
\noindent\reffig{fig:pseudo_ens} shows the pseudo-labels obtained by averaging the predictions of two target networks $F_{t,\sigma^2=1}$, $F_{t,\sigma^2=10}$ and a one source network $F_s$. Averaging the predictions allows us to generate more accurate pseudo-labels.

\begin{figure*}[h]
\centering
    \includegraphics[width=\textwidth]{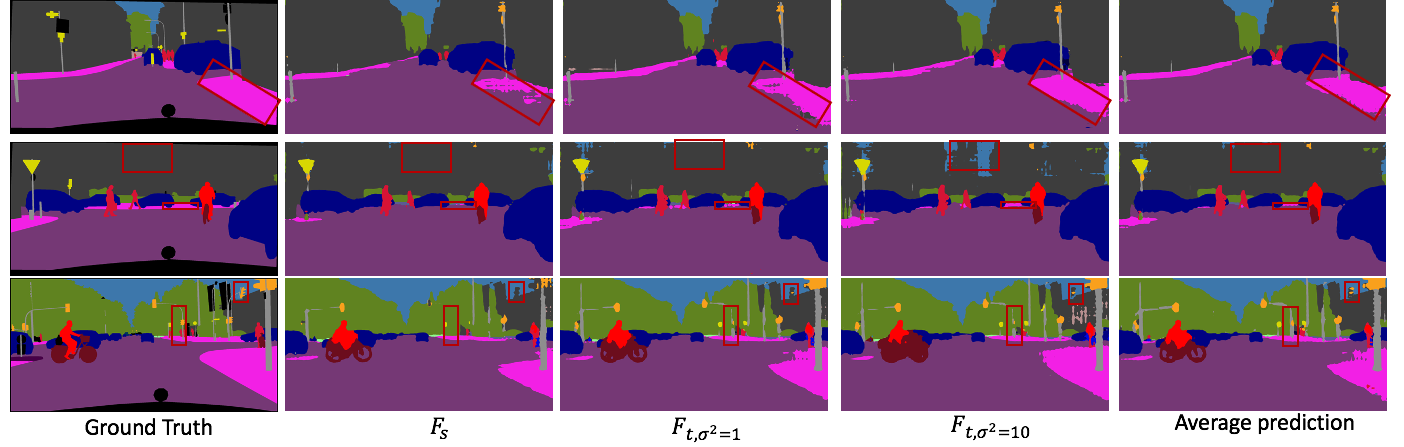}
\caption{Visualization of results obtained by averaging the predictions of two target networks $F_{t,\sigma^2=1}$, $F_{t,\sigma^2=10}$ and a one source network $F_s$. Averaging the predictions allows us to further improve performance by better distinguishing similar structures (e.g., road, sidewalk) and  identifying small objects.}
\label{fig:ens}
\end{figure*}

\noindent\textbf{Ensembling for improved segmentation performance}. \\
\noindent\reffig{fig:ens} shows the predictions obtained by averaging the predictions of two target networks $F_{t,\sigma^2=1}$, $F_{t,\sigma^2=10}$ and a one source network $F_s$. Averaging the predictions allows us to further improve performance by better distinguishing similar structures (e.g., road, sidewalk) and  identifying small objects.

\begin{figure*}[h]
\centering
    \includegraphics[width=\textwidth]{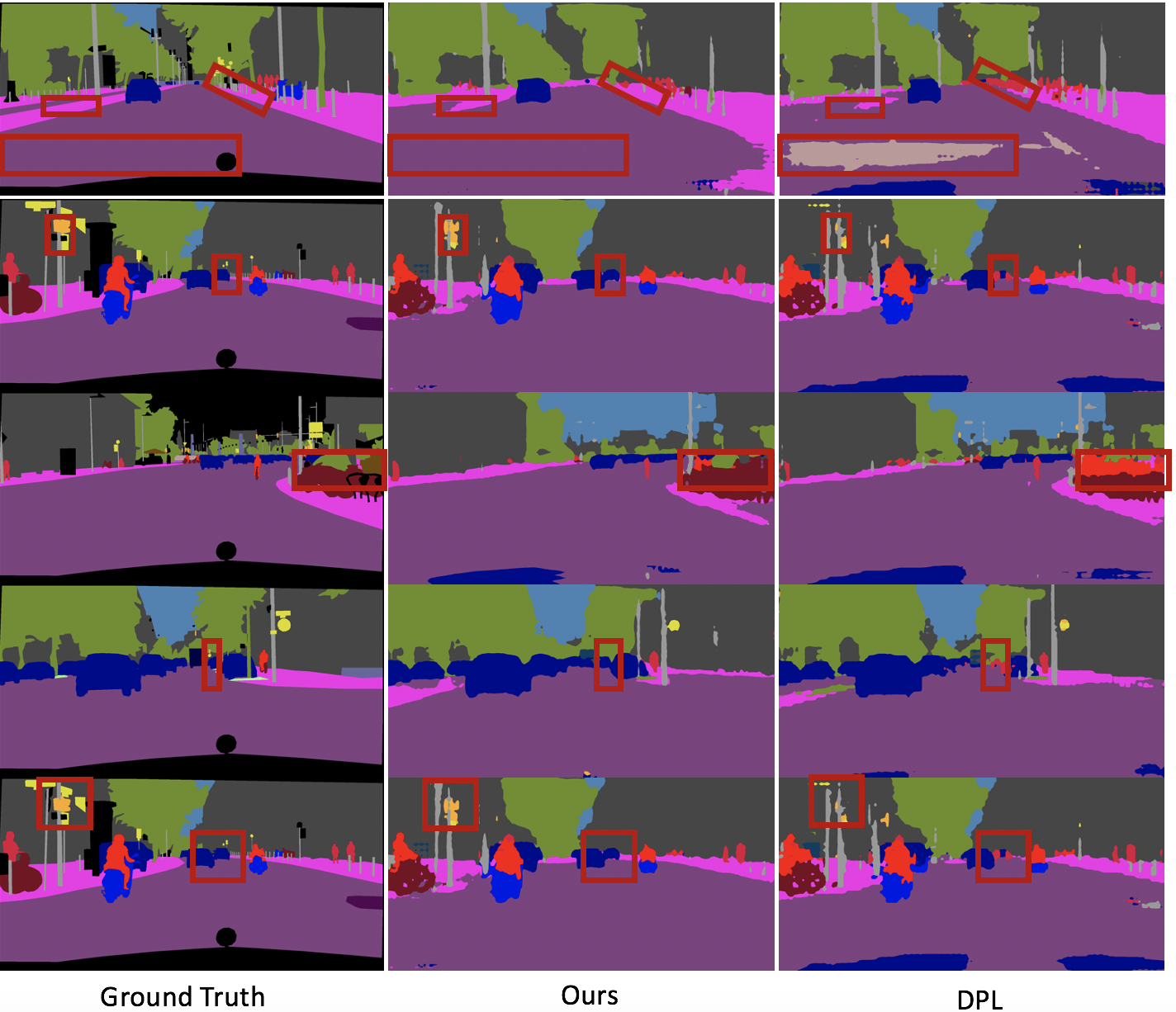}
\caption{Qualitative comparison of our method with DPL\cite{Cheng_21_dual_path}. Our method generates better predictions that are closer to the ground-truth.}
\label{fig:qual_comp}
\end{figure*}

\noindent\textbf{Qualitative comparison of the segmentation results}. \\
\noindent\reffig{fig:qual_comp} shows results segmentation results obtained by our method and DPL\cite{Cheng_21_dual_path}. Our method generates better predictions that are closer to the ground-truth. 
\end{document}